\documentclass[10pt,journal,compsoc]{IEEEtran}

\ifCLASSOPTIONcompsoc
  \usepackage[nocompress]{cite}
\else
  \usepackage{cite}
\fi
\ifCLASSINFOpdf
  \usepackage[pdftex]{graphicx}
\else
  \usepackage[dvips]{graphicx}
\fi
\usepackage{amsmath}
\usepackage{array}
\usepackage{graphicx} 
\usepackage{algorithm}
\usepackage{algpseudocode}
\usepackage{amsmath}
\usepackage{amsfonts}
\usepackage{amssymb}
\usepackage{balance}
\usepackage{booktabs} 
\usepackage[dvipsnames]{xcolor}
\begin{document}
%
\title{Attention-aware Social Graph Transformer Networks for Stochastic Trajectory Prediction}
%
%
%
%

\author{Yao Liu,
        Binghao Li, 
        Xianzhi Wang,
        Claude Sammut,
        Lina Yao,~\IEEEmembership{Senior Member,~IEEE}
\IEEEcompsocitemizethanks{\IEEEcompsocthanksitem Yao Liu is with the School of Computer Science and Engineering, the University of New South Wales, Sydney, NSW 2052, Australia.\protect\\
E-mail: yao.liu3@unsw.edu.au}
\IEEEcompsocitemizethanks{\IEEEcompsocthanksitem Binghao Li is with the School of Minerals and Energy Resources Engineering, the University of New South Wales, Sydney, NSW 2052, Australia.\protect\\
E-mail: binghao.li@unsw.edu.au}
\IEEEcompsocitemizethanks{\IEEEcompsocthanksitem Xianzhi Wang is with the School of Computer Science, the University of Technology Sydney, Sydney, NSW 2007, Australia.\protect\\
E-mail: xianzhi.wang@uts.edu.au}
\IEEEcompsocitemizethanks{\IEEEcompsocthanksitem Claude Sammut is with the School of Computer Science and Engineering, the University of New South Wales, Sydney, NSW 2052, Australia.\protect\\
E-mail: c.sammut@unsw.edu.au}
\IEEEcompsocitemizethanks{\IEEEcompsocthanksitem Lina Yao is with the Data 61, CSIRO and the School of Computer Science and Engineering, the University of New South Wales, Sydney, NSW 2015, Australia.\protect\\
E-mail: lina.yao@unsw.edu.au}
\thanks{Manuscript received November **, 2022}}

\IEEEtitleabstractindextext{%
\begin{abstract}
Trajectory prediction is fundamental to various intelligent technologies, such as autonomous driving and robotics.
The motion prediction of pedestrians and vehicles helps emergency braking, reduces collisions, and improves traffic safety.
Current trajectory prediction research faces problems of complex social interactions, high dynamics and multi-modality. Especially, it still has limitations in long-time prediction.
We propose Attention-aware Social Graph Transformer Networks for multi-modal trajectory prediction. We combine Graph Convolutional Networks and Transformer Networks by generating stable resolution pseudo-images from Spatio-temporal graphs through a designed stacking and interception method.
{
Furthermore, we design the attention-aware module to handle social interaction information in scenarios involving mixed pedestrian-vehicle traffic.
}
Thus, we maintain the advantages of the Graph and Transformer, i.e., the ability to aggregate information over an arbitrary number of neighbors and the ability to perform complex time-dependent data processing.
{
We conduct experiments on datasets involving pedestrian, vehicle, and mixed trajectories, respectively. Our results demonstrate that our model minimizes displacement errors across various metrics and significantly reduces the likelihood of collisions.}
It is worth noting that our model effectively reduces the final displacement error, illustrating the ability of our model to predict for a long time.
\end{abstract}

}

\maketitle

\IEEEdisplaynontitleabstractindextext

%
\IEEEpeerreviewmaketitle


\section{Introduction}
\IEEEPARstart{T}{raffic} prediction is crucial for building smart cities and analyzing the modern society.
For instance, traffic flow forecasting helps reduce congestion and optimize road construction~\cite{flow2, flow3}, while destination forecasting assists in analyzing gathering events and enhancing public safety~\cite{event1, event2}. 
However, motion trajectory prediction is more challenging, as it requires predicting future locations based on the current and past locations of traffic agents~\cite{Human_survey, TrafficPredict}. 
Traffic agents primarily encompass pedestrians and vehicles.
In the context of autonomous driving and robotic navigation, accurate motion prediction is vital to avoid collisions and implement emergency braking~\cite{Crowd-Robot_Interaction, DESIRE}. In underground mining scenarios, vehicles can pose a significant threat to human safety in tight and low-light conditions~\cite{YOLOv5, Autonomous}.
Humans can naturally navigate complex areas with social interactions because they can easily observe other agents and objects, recognize signs, and make route judgments based on common sense and social conventions.
An agent's moving route exhibits both certainty and randomness.
Certainty is evident in pedestrians walking in parallel with others in the same direction and avoiding collisions with those moving in the opposite direction. Drivers adjust their speed according to road signs and the presence of surrounding vehicles.
However, randomness emerges from the fact that the motivation for human movement originates from humans themselves. For prediction, the expected future trajectory of a person's movement is inherently random, and the destination is often unknown.

\begin{figure}[t]
  \centering
  \includegraphics[width=3.5in]{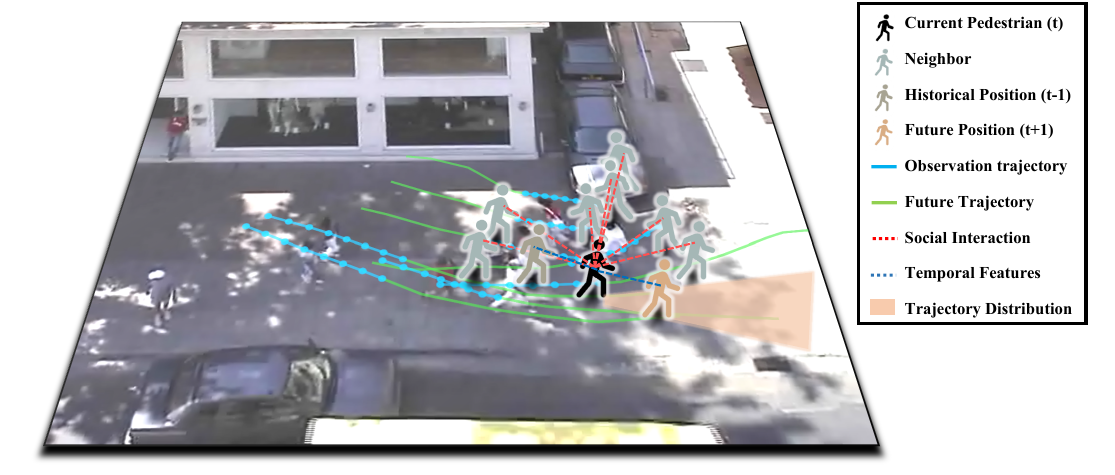}
  \caption{An example of pedestrian motion trajectory in the real scenario. Pedestrian trajectory prediction needs to consider social interaction information, and pedestrian trajectories are dynamic and multi-modal.}
  \label{figure-real_scenario}
\end{figure}

\IEEEpubidadjcol

Motion trajectory prediction for complex social scenarios presents several challenges because machines cannot navigate based on experience and habits like humans. 
The first challenge is related to social interaction information. When humans walk, they must take into account the people around them and make judgments~\cite{Social_LSTM, Trajectron++}. For instance, in a group, humans often follow the pace of their peers, and individuals may slow down or even stop to avoid collisions.
Similarly, drivers need to adjust their speed based on the vehicles in front of them and be aware of vehicles behind them when turning or changing lanes~\cite{GRIP, cslstm}.
The second challenge lies in the dynamic nature of the environment in motion trajectory prediction~\cite{TrafficPredict, S2TNet, Trajectron}. In a given area of interest, agents can enter or exit at any time, resulting in a varying number of agents in that area as time progresses.
The third challenge pertains to multi-modal trajectory prediction of motions~\cite{Social_LSTM, Social_GAN}.
Multi-modality implies that the future trajectory of motion encompasses numerous possibilities, and humans may follow one of several plausible trajectories from the potential distribution. 
While social interactions influence agents' movements, their ultimate destination remains subject to individual decisions. Therefore, trajectory prediction aims to forecast the probability distribution of an agent's future trajectory.
Figure~\ref{figure-real_scenario} illustrates an example of human motion in a real scenario, highlighting the complexity of predicting pedestrian trajectories in such socially intricate situations. 
We are motivated by these challenges and seek to design a model that can not only comprehend social interaction information and predict trajectories but also dynamically adapt to varying numbers of agents and perform multi-modal trajectory prediction over extended time periods.

An early approach that incorporates social information into trajectory prediction is the Social Force model~\cite{Social_force}. This model represents human motion as a dynamic system and predicts human trajectories by considering both attractive forces between individuals and their targets, as well as repulsive forces arising from interactions with other people.
Another approach, introduced in \cite{Gaussian_Process}, treats trajectory prediction as a sequential data regression problem and employs Gaussian processes for regression modeling. However, it's worth noting that Gaussian process regression can be computationally intensive and lead to long inference times.
With the advancement of deep learning, Recurrent Neural Networks (RNNs) and Long Short-Term Memory networks (LSTMs)~\cite{LSTM} have gained widespread use for processing sequential data and have demonstrated advantages over traditional methods. In RNN-based methods, models capture motion in the temporal dimension using hidden states while incorporating social interaction information by merging the feature states of agents in the spatial dimension.
However, it's important to acknowledge that LSTM-based methods, as highlighted in \cite{Attention_is_All}, may struggle to adequately handle complex time-dependent relationships, as they rely on a single vector to record sequence history with limited memory capacity.
More recently, the Transformer architecture, which achieved breakthroughs in Natural Language Processing (NLP)\cite{Attention_is_All, BERT}, has also shown promise in processing time-series data. Transformers leverage a self-attention mechanism for temporal modeling, improving modeling effectiveness and enabling parallel computation. This adaptability makes Transformers well-suited for addressing trajectory prediction tasks\cite{flow1}.

{
Building upon prior research~\cite{yaoliu}, we propose a novel model, the Attention-aware Social Graph Transformer Networks, for predicting motion trajectories in complex real-world public areas. Our model seamlessly integrates the capabilities of the Graph Convolutional Network (GCN) and the Transformer Network through a carefully designed stacking and interception method.
While GCN excels at handling varying amounts of neighbor information, its output tends to fluctuate with changes in input, lacking stability. In contrast, the Transformer Network is proficient at processing long and intricate time-dependent data but typically requires a consistent input. In Transformer Networks, dealing with extreme values using the fill-in approach can significantly increase computational costs, and simple interceptions may result in the loss of critical information, with the number of interceptions often relying on subjective empirical judgments. The challenge lies in effectively combining GCN and Transformer Networks.
Our model achieves stable integration of both networks while preserving their respective advantages through the introduction of pseudo-images. These pseudo-images gather diverse neighbor information via the Social Spatio-temporal Graph Convolutional Network (SSTGCN) and generate stable-resolution inputs for Transformer-based spatio-temporal predictions. 
Furthermore, we introduce an innovative Attention mechanism to our SSTGCN, enabling a more rational collection of social information. Unlike pedestrians, vehicles on the road move faster and in larger numbers. When controlling vehicles, drivers typically focus their attention on specific areas requiring immediate attention, rather than monitoring all surrounding vehicles. Consequently, the Attention-aware and Sparse Attention-aware SSTGCN can better model driver attention.
We apply this model to multi-modal trajectory prediction.
To the best of our knowledge, our model represents the first attempt to capture spatio-temporal features through GCN and make predictions using Transformer networks. While a previous model~\cite{STAR} combines Graph and Transformer elements, it employs `graph' as the structural framework for implementing the Transformer, fundamentally differing from our approach that directly leverages an adjacency matrix to capture spatio-temporal features through GCN.
}

{
In brief, our main contributions are as follows: 
\begin{itemize}
\item We propose Attention-aware Social Graph Transformer Networks, which retain the advantages of Graph Convolutional Networks and Transformer Networks without introducing any additional information, to predict the trajectory by coordinates.
\item Our model constructs trajectories as Social Interaction Spatio-temporal Graphs and converts them into pseudo-images using a designed stacking and interception method. The constant resolution pseudo-images capture Spatio-temporal information by Graph Convolutional Networks and complete temporal extrapolation by Transformer Networks. Our proposed Attention-aware and Sparse Attention-aware modules demonstrate their advantages in vehicle trajectory prediction, as well as in predicting trajectories in mixed-traffic scenarios. They allow for a more rational collection of social interaction information compared to pedestrian trajectory prediction.
\item We perform multi-modal trajectory prediction for pedestrians and vehicles as well as for mixed traffic agents.
Multi-modal predictions on the benchmark show that our model reduces the displacement error and achieves state-of-the-art in most metrics.
In particular, the reduction in the final displacement error indicates our model's ability to make a long-time prediction.
Additionally, we also assess the collision rate in the pedestrian trajectory prediction results, and our findings indicate a significant reduction in the likelihood of collisions. The ablation experiments validate the contribution of each module, and the visualization visually compares the advantages of our model relative to other models.
\end{itemize}
}
\section{Related Work} 

\subsection{RNN-based Methods}
Recurrent Neural Networks (RNNs), including Long Short-Term Memory networks (LSTMs)~\cite{LSTM}, have achieved significant success in the field of trajectory prediction.
One of the pioneering methods to integrate social interaction information with LSTM is Social LSTM~\cite{Social_LSTM}. Social LSTM introduces a pooling mechanism that aggregates feature information around the agent in space and utilizes LSTM to output the predicted trajectory. However, this aggregation approach is inefficient, as it treats neighbors in each grid equally, making it challenging to obtain accurate features. Furthermore, Social LSTM introduces the concept that pedestrian trajectory prediction follows a bivariate Gaussian distribution, setting an early example for multi-modal trajectory prediction.
Subsequently, Social GAN~\cite{Social_GAN} was introduced to address the multi-modality of pedestrian trajectories. It combines a novel pooling mechanism with a Generative Adversarial Network (GAN) model. SR-LSTM~\cite{SR-LSTM} takes a different approach, using a message-passing and selection model to capture the key contributions of neighbors and improves Social LSTM by incorporating visual features and a new pooling mechanism.
SoPhie~\cite{SoPhie} employs Convolutional Neural Networks (CNNs) to extract scene information, followed by a two-way attention mechanism to extract social features. Finally, LSTMs are used to generate future trajectories. CS-LSTM~\cite{cslstm} introduces convolutional social pooling for gathering surrounding feature information, performs vehicle trajectory prediction using LSTM, and conducts multi-modal prediction based on maneuver classes. TrafficPredict~\cite{TrafficPredict} models social interactions among agents through soft attention and leverages LSTM to capture the similarity of motions within the same category for prediction.
It's worth noting that many approaches based on RNNs adopt a two-step structure: first, they aggregate social interaction information through pooling or attention mechanisms, and then they make predictions using networks like LSTMs. However, methods relying on LSTMs to record history using a single vector with limited memory may face challenges when dealing with sequential data characterized by complex time dependencies.

\subsection{Transformer-based Methods }
Transformers, as exemplified by the work in \cite{Attention_is_All}, are well-suited for modeling sequential data and trajectory prediction, particularly for dealing with nonlinear patterns and datasets characterized by complex temporal relationships. The key to the success of Transformers lies in their self-attention mechanism, which enables them to consider the entire sequence while assigning varying levels of importance to different elements, ultimately enhancing the quality of feature learning.
In the context of trajectory prediction, \cite{Transformer_for_Trajectory} demonstrates promising results by directly employing the self-attention mechanism for pedestrian trajectory prediction, without explicitly considering social interaction or scene information. Interaction Transformer~\cite{End-to-end_Transformer} takes a different approach by explicitly incorporating social interactions among agents in autonomous driving environments, combining a Recurrent Neural Network with a Transformer structure.
mmTransformer~\cite{mmTransformer} adopts a stacked transformer as the core component for trajectory prediction, aggregating historical trajectories, social interactions, and road map information. S2TNet~\cite{S2TNet}, on the other hand, builds upon the basic Transformer architecture. It utilizes self-attention to capture social interaction information among agents in the spatial dimension and employs a temporal convolutional network to extract temporal feature dependencies. The model then regresses future trajectories through a temporal self-attention mechanism. S2TNet's historical feature information encompasses various agent attributes, such as coordinates, length and width, direction, and category, making it suitable for predicting traffic trajectories in mixed pedestrian-vehicle scenarios.
Despite the significant improvements that Transformer-based methods offer over LSTM-based methods for sequence data prediction, these methods may still face challenges when predicting the trajectories of agents in complex areas.

\subsection{Graph-based Methods}
The graph structure is naturally suited for aggregating feature information from surrounding neighbors. By representing agents as vertices and social interaction information between agents as edges, the graph structure can easily handle feature aggregation with varying numbers of neighbors.
Social-BiGAT~\cite{Social-BiGAT} combines the graph structure with Generative Adversarial Networks (GANs) to introduce a graph attention network for learning people's social interactions in a scene. This network is then used for multi-modal prediction of pedestrian trajectories through recurrent networks.
Similarly, GRIP~\cite{GRIP} models social interactions among closely located agents using graphs, extracts aggregated features through graph convolution, and employs LSTM models for highway vehicle trajectory prediction.
TPCN~\cite{TPCN} performs trajectory prediction by treating agent coordinates as point clouds. While it does not explicitly consider kinematic and motion information, it introduces dynamic temporal learning to model motion and directly uses the point cloud approach for prediction.
Social-STGCNN~\cite{Social-STGCNN} introduces a Social Spatio-Temporal Graph Convolutional Neural Network that constructs a graph structure based on social interactions between agents. It leverages the adjacency matrix instead of aggregation methods to acquire social information. Trajectory prediction is performed using a convolutional structure, and the model structure does not include any recurrent networks or Transformer structures.
Our model draws inspiration from ST-GCNN~\cite{ST-GCNN} and Social-STGCNN. Initially, Spatio-temporal Graph Convolutional Networks were applied in skeletal action recognition, and later they were introduced for traffic prediction.
STAR~\cite{STAR} introduces a Spatio-temporal Graph within the Transformer structure for pedestrian trajectory prediction. In STAR, for the temporal dimension, the standard Transformer is used to extract features independently for each pedestrian. For the spatial dimension, graph convolution is performed through the message-passing mechanism of the Transformer to capture social interaction information among pedestrians. Features are then merged using a fully connected layer to obtain Spatio-temporal information.
We argue that in STAR, the graph primarily serves as a structural framework for obtaining social interaction information through the Transformer. Subsequently, the merged Spatio-temporal information is acquired through a fully connected network. This approach fundamentally differs from our model, where Spatio-temporal features are directly obtained through convolutional operations using the Spatio-temporal Graph and the adjacency matrix.

\section{Methodology}

\subsection{Problem Formulation and Model Overview}

\begin{figure*}[t]
  \centering
  \includegraphics[width=6.5in]{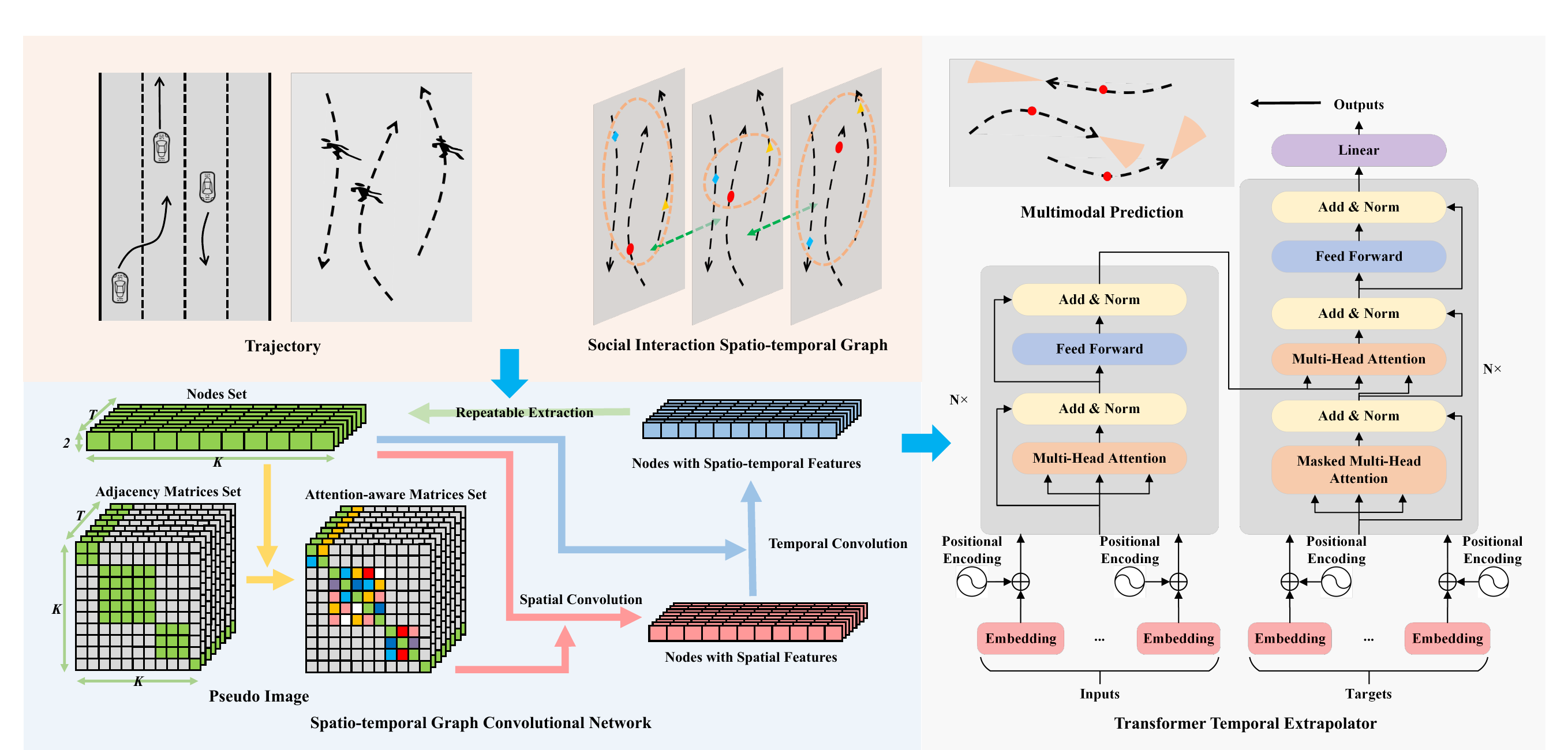}
  \caption{{The overview of our Attention-aware Social Graph Transformer Networks. It mainly consists of Social Interaction Spatio-temporal Graph, Spatio-temporal Graph Convolutional Network with pseudo-image and Attention-aware module, Transformer Temporal Extrapolator, and multi-modal prediction.}}
  \label{figure-overview}
\end{figure*}

Our goal is to predict trajectories within public areas under complex social interaction scenarios. 
The dataset we are working with provides solely the positional information of agents, without any additional scene or map data, and without details such as length, width, or direction. Each agent is represented as a point.
Our problem is defined as follows: Given an observed trajectory, denoted as $Traj_{obs}$, which contains the locations of all agents from timestamp 1 to $T_{obs}$, with the possibility of a varying number of agents at each timestamp, we aim to predict the trajectory $Traj_{pred}$ of agents from $T_{obs}+1$ to $T_{obs}+T_{pred}$.

\begin{align}
    \begin{aligned}
        Traj_{obs}= \{   
        p_{t}^{n}&=(x_{t}^{n},y_{t}^{n})~|~ 
        n=f(t) \in   \{ 1, \cdots, N \},\\
        t & \in   \{ 1, \cdots, T_{obs}  \}
        \}
    \end{aligned}
\end{align}

\begin{align}
    \begin{aligned}
        Traj_{pred}= \{   
        \hat{p}_{t}^{n}&=(\hat{x}_{t}^{n},\hat{y}_{t}^{n}) ~|~
        n=f(t) \in   \{ 1, \cdots, N  \},\\
        t & \in   \{ T_{obs} + 1, \cdots, T_{obs}+T_{pred}  \}
         \}
    \end{aligned}
\end{align}

We approach trajectory prediction as a multi-modal problem, where each agent can select one of the plausible future trajectories. Our predicted future trajectories are represented as probability distributions encompassing these plausible trajectories.
Our objective is to ensure that the ground truth trajectory, denoted as $Traj_{gt}$, falls within a plausible sample drawn from the distribution of $Traj_{pred}$.

Our model is shown in Figure~\ref{figure-overview}.
Social Interaction Spatio-temporal Graph (SSTG) and pseudo-image can organize discrete points in space and time into a graph structure and form a compact representation through the adjacency matrix so that Spatio-temporal features can be extracted using a convolution-like approach.
{
To assign weights to the adjacency matrix in a more informed manner, we introduce an attention-aware module. This module incorporates both attention mechanisms and sparse attention mechanisms. By doing so, we aim to improve the efficiency of aggregating social interaction information while reducing redundancy.
}
Like the Social-STGCNN~\cite{Social-STGCNN}, our SSTG can also accept a variable number of trajectories. 
However, We make a significant improvement by outputing features with constant resolution (the number of agents) through the designed adjacency matrix. 
The constant resolution helps enhance the stability of the model and improve the performance, which is well-proven in CNN methods~\cite{RCNN}, and it is why numerous pooling methods are proposed. 
The constant resolution output serves as a solid foundation for the subsequent application of the Transformer model.
We use the Transformer as a temporal extrapolator to predict future trajectories from Spatio-temporal features during the observation time. 
The Transformer can encode timestamps and use parallel computation through positional encoding and mask. 
In this way, we do not need to perform data preprocessing, and the model can be unaffected by the missing data of some agents in some timestamps.
Finally, the multi-modal mechanism outputs distribution intervals of plausible future trajectories through bivariate Gaussian distributions.

\subsection{Social Interaction Pseudo-image}

First, we define the structure of our graph.
We denote each traffic participant as an agent, and for a timestamp $t$, we form a graph structure $G_{t}$ with all $N$ agents.
$G_{t} = (V_{t}, E_{t})$, where $V_{t}$ is the vertex set, and each agent is a vertex; $E_{t}$ is the edge set, which represents the social interaction between two agents.
$V_{t}$ contains the coordinates of the agents,  and $E_{t}$ is represented as the distance between the agents measured by the L2 norm.
\begin{equation}
    V_{t}=\{ v_{t}^{n}=(x_{t}^{n},y_{t}^{n})~|~ \forall n \in (1,\cdots,N) \}
\end{equation}
\begin{equation}
     E_{t}=\{e_{t}^{n,m}=||v_{t}^{n}-v_{t}^{m}||_{2}~|~ \forall n,m \in (1,\cdots,N) \}
\end{equation}

\begin{figure}[t]
  \centering
  \includegraphics[width=0.7\linewidth]{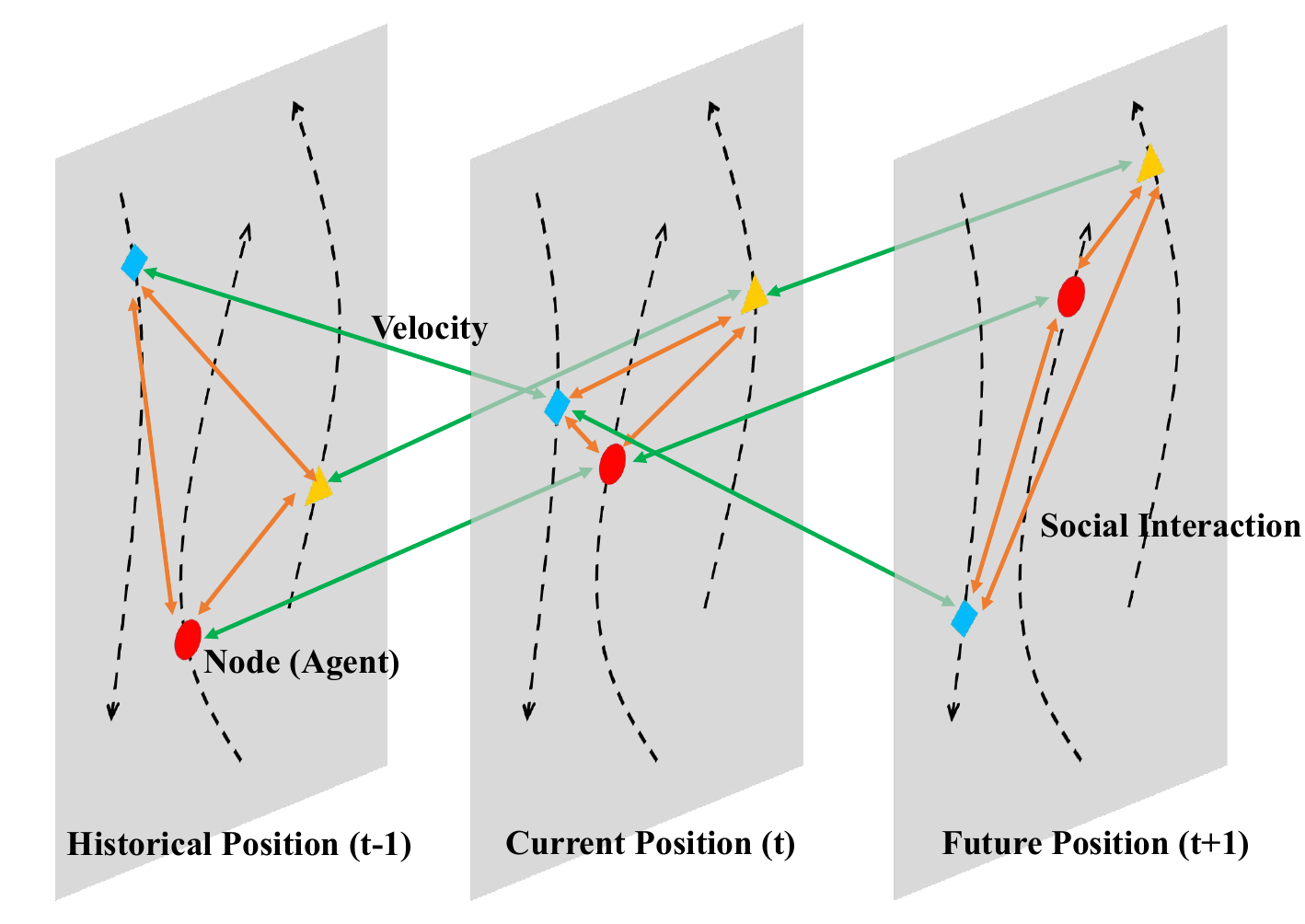}
  \caption{Social Interaction Spatio-temporal Graph.}
  \label{figure-graph_structure}
\end{figure}

After that, we perform Spatio-temporal expansion on the graph $G_{t}$ to form the Spatio-temporal graph ${G_{t}}' = ({V_{t}}', {A_{t}}')$ (shown in Figure~\ref{figure-graph_structure}),
where the node set ${V_{t}}'$ is the displacement per unit time, known as `velocity,' as a way to represent the temporal features while stabilizing the model input; ${A_{t}}'$ is the adjacency matrix, generated by the edge set $E_{t}$, which represents the social interaction between nodes.

\begin{equation}
{V_{t}}'= \begin{Bmatrix}
(0,0)^{n} &|& t=1\\ 
\Delta v_{t}^{n} &|& t \in   \{ 2, \cdots, T_{obs} \}
\end{Bmatrix}
\end{equation}
\begin{equation}
A_{t}=\{a_{t}^{n,m}=f_{a}(e_{t}^{n,m})~|~ \forall n,m \in (1,\cdots,N)\}
\end{equation}
\begin{equation}
{A_{t}}'=\Lambda_{t}^{\frac{1}{2}}A_{t}\Lambda_{t}^{-\frac{1}{2}}
\end{equation}

In the node set ${V_{t}}'$, we assume that the velocity of all agents in the first time step is 0, and the subsequent velocities are obtained from the position information with timestamps.
The adjacency matrix $A_{t}$ is obtained by computing the L2 distance $e_{t}^{n,m}$ between nodes with the kernel function $f_{a}$, which represents the social interaction factor between each node.
$ \Lambda_{t} $ is the diagonal nodal degree matrix of $A_{t}$ and ${A_{t}}'$ is the normalized matrix of $A_{t}$.
The normalized adjacency matrix ${A_{t}}'$ is obtained by symmetrically normalizing $A_{t}$ with $ \Lambda_{t} $, which is a critical treatment for the proper computation of the graph convolution~\cite{graph-cnn}.
We choose the inverse of the L2 distance as the kernel function $f_{a}$.

\begin{equation}
f_{a}=\begin{Bmatrix}
1 &|& L2_{dist} = 0 \\ 
\frac{1}{L2_{dist}}&|& L2_{dist} \neq  0
\end{Bmatrix}
\end{equation}

Finally, we obtain the pseudo-image $\tilde{G}$ by stacking the Spatio-temporal graph ${G_{t}}'$. In our pseudo-image $\tilde{G}$, both the original graph structure is maintained, and a constant resolution is guaranteed, which allows the use of graph convolution operations to obtain Spatio-temporal features and also helps to combine with the Transformer module later.

\begin{equation}
\tilde{G}=\{ \tilde{V} ,\tilde{A} \}
\end{equation}
\begin{equation}
\tilde{V}=recombine({V}')\in \mathbb{R}^{T\times K \times 2}
\end{equation}
\begin{equation}
\tilde{A}=recombine({A}')\in \mathbb{R}^{T\times K \times K} 
\end{equation}

\begin{figure}[t]
  \centering
  \includegraphics[width=3.5in]{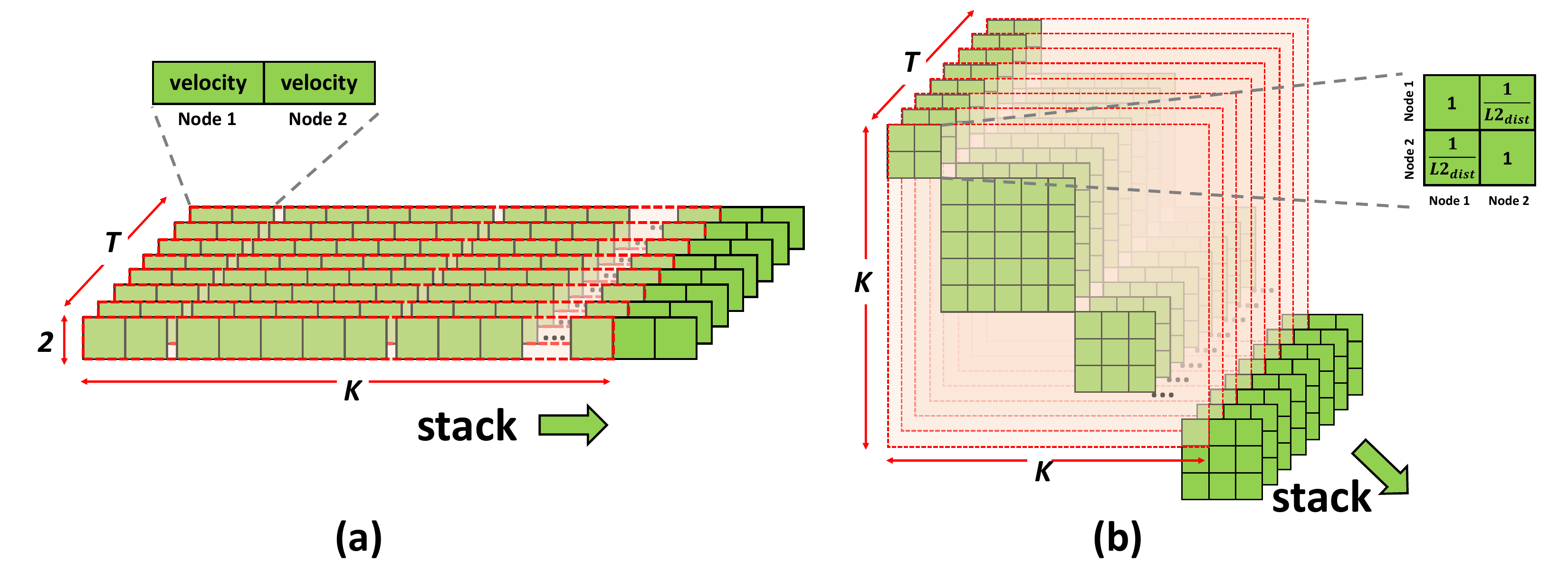}
  \caption{{Pseudo-image. The node set (a) and adjacency matrix set (b) of the Spatio-temporal graph are stacked and intercepted in a certain way to obtain constant resolution node set pseudo-images and adjacency matrix pseudo-images.}}
  \label{figure-pseudo_image}
\end{figure}

The resolution here refers to the length and width dimensions in similar images. By recombining ${G}'$, we can compose pseudo-images $\tilde{G}$ from graphs containing varying numbers of nodes.
For the node set ${V}'$, first, the nodes are arranged in time, and the time steps $T_{obs}$ are fixed for the same task; then, the nodes are stacked, and a fixed number of nodes ($K$) are intercepted each time; each cell is node features that are velocities.
For the set of adjacency matrices ${A}'$, the first dimension is arranged by time, corresponding to the set of nodes; then, in each time step, the matrices are stacked diagonally and intercepted with a square matrix of size $K^{2}$; each cell is the weight parameter of the adjacency matrix, and the empty places are filled with 0, as these nodes do not have social interactions. The stacking and interception are shown in Figure~\ref{figure-pseudo_image}.

\subsection{Attention-aware Adjacency Matrix}
\begin{figure}[t]
  \centering
  \includegraphics[width=3in]{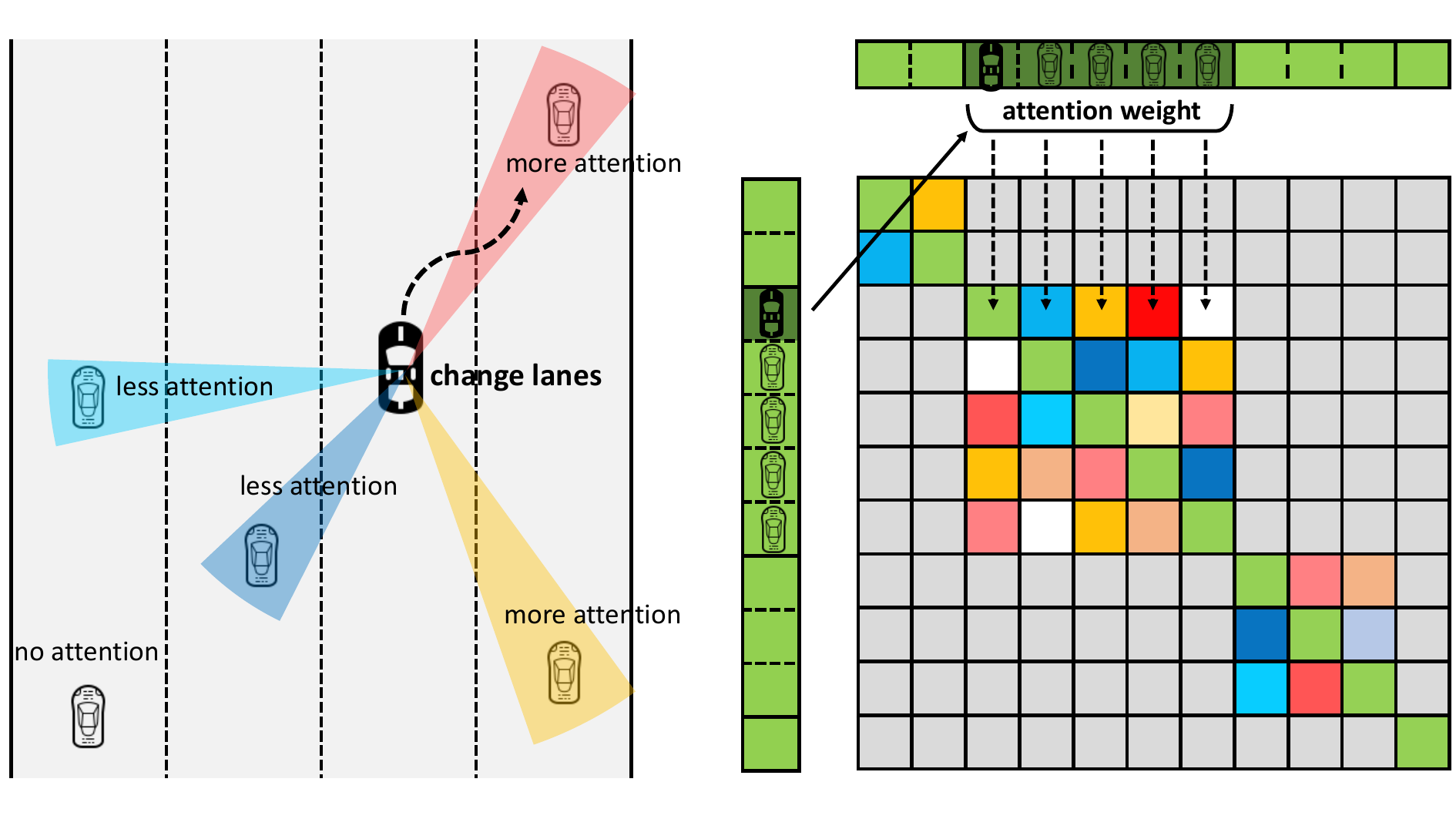}
  \caption{{Attention-aware adjacency matrix. When drivers switch roads in their cars, their attention is focused on the vehicles in front and behind them on this side and less on the vehicles on the other side.}}
  \label{figure-att}
\end{figure}

{
The characteristics of vehicle trajectories differ from those of pedestrian trajectories.
Vehicles on the road are faster and more numerous than pedestrians. Our Social Interaction Spatio-temporal Graph defines the influence of surrounding agents through the adjacency matrix. 
Pedestrians continuously adjust their walking direction by observing the people around them. Typically, pedestrians move at lower speeds and often navigate through crowded spaces, which necessitates constant awareness of their surroundings.
In the case of on-road vehicles, drivers typically need to maintain a higher level of focus. This means that they often concentrate on specific areas of the road, especially when they are moving at high speeds.
When driving in a straight line, they need to pay attention to the vehicles in front of them, as well as the vehicles on either side that need to merge, and are less likely to pay attention to the vehicles behind them; when turning and switching lanes, they will pay attention to the vehicles in front and behind them in this direction, and less likely to pay attention to the vehicles on the other side, as shown in Figure~\ref{figure-att}.
Hence, it is necessary to apply suitable extensions of the adjacency matrix based on the distance metric to predict trajectories for different types of traffic agents.
The attention mechanism provides good practice for determining weights by assigning queries and keys to values.
}

{
Attention provides a flexible and global mechanism to handle dynamic and long-time trajectory data more rationally and effectively~\cite{flow4}.
We make the adjacency matrix expand into an Attention-aware adjacency matrix by introducting an Attention-aware mechanism to the pseudo-image.
Specifically, the attention weights are assigned to the adjacency matrix by the adjacency matrix itself. Then element-wise multiplication is performed to obtain the new Attention-aware adjacency matrix.
In addition, we also design the Sparse Attention-aware adjacency matrix. After obtaining the attention weights, a threshold determines whether the attention should be retained or not, which is then multiplied by the elements of the adjacency matrix.
In practice, we set the threshold to 0.5, and in our experiments, we will compare the prediction results of various different types of traffic agents with and without the attention-aware module.
}

{
\begin{equation}
attn=softmax(Attention~Score(Q,K))
\end{equation}
\begin{equation}
sp\text{-}attn=\left\{\begin{matrix}
1 &|& attn>0.5\\ 
0 &|& attn\leqslant 0.5
\end{matrix}\right.
\end{equation}
\begin{equation}
attn\text{-}\tilde{A}=attn~\odot~\tilde{A}
\end{equation}
\begin{equation}
sp\text{-}attn\text{-}\tilde{A}=sp\text{-}attn~\odot~\tilde{A}
\end{equation}
}
{
In this context, `$attn$' represents the attention weight, `$sp\text{-}attn$' represents the sparse attention weight, and `$Q$' and `$K$' refer to the query and key, respectively, specifically related to the encoding and decoding of the adjacency matrix `$\tilde{A}$' itself.
Please note that in the subsequent sections describing the methodology, we will continue to use the representation `$\tilde{A}$' for the adjacency matrix for the sake of clarity in our explanations.
}

\subsection{Spatio-temporal Graph Convolutional Network}
Our Social Interaction Spatio-temporal Graph Convolutional Network consists of spatial convolution and temporal convolution.
Since we obtain the pseudo-image $\tilde{G}$ by graph $G$, our data organization is dense, so we can easily use the convolution process.
{
For spatial convolution $Conv_{spa}$, we employ a two-step process that involves a  Convolutional Neural Networks (CNNs) and adjacency matrices $\tilde{A}$.
In the convolutional neural network, we use a kernel of $(H_{s}, W_{s}) = (1,1)$, without padding, and a 2-dimensional feature of the input $\tilde{V}$ with a 5-dimensional feature of the output, which is used to expand the dimensions and facilitate the multi-modal prediction afterward.
The input and output resolutions of the CNN remain constant. The social interaction information for each node is derived from the adjacency matrices. Social interaction features are computed by matrix multiplication and summation of the feature matrices and adjacency matrices, all while ensuring that the output resolution remains constant.
}
The temporal convolution $Conv_{temp}$ is also composed of two parts, the convolution part is mainly responsible for aggregating temporal features, and the residual part ensures the model's stability.
The convolution part uses kernel $(H_{t}, W_{t}) = (3,1)$ while paddings are added, and the input and output are kept constant in both resolution and feature dimensions.
The kernel of $(H_{t}, W_{t}) = (3,1)$ indicates that each node aggregates the features from one time step before and after the current time point.
The residual network expands the feature dimension by the convolutional network at the first layer and uses node features directly afterward, which highlights the features at the current time point and stabilizes the model.
After our Social Interaction Spatio-temporal Graph Convolution, $\tilde{V}$ keeps the resolution constant, and each node at each time step aggregates the information of nodes with social interactions in the surroundings as well as the features of the one time step before and after.

\begin{equation}
\tilde{V}^{l+1}=Conv_{temp}(Conv_{spa}(\tilde{G}^{l}))
\end{equation}
\begin{equation}
Conv_{spa}(\tilde{G}^{l}) = ( \sum_{h=1}^{H_{s}} \sum_{w=1}^{W_{s}} \tilde{V}^{l} \cdot \mathbf{W_{s}}^{l}(h,w) + bias )\circledast \tilde{A}
\end{equation}
\begin{equation}
\tilde{Z}^{l} = Conv_{spa}(\tilde{G}^{l})
\end{equation}
\begin{equation}
Conv_{temp}(\tilde{Z}^{l},\tilde{V}^{l}) = ( \sum_{h=1}^{H_{t}} \sum_{w=1}^{W_{t}} \tilde{Z}^{l} \cdot \mathbf{W_{t}}^{l}(h,w) + bias ) +Res(\tilde{V}^{l})
\end{equation}
\begin{equation}
Res(\tilde{V}^{l})=\left\{\begin{matrix}
\sum_{h=1}^{H_{r}} \sum_{w=1}^{W_{r}} \tilde{V}^{l} \cdot \mathbf{W_{r}}^{l}(h,w) &|& l=1\\ 
\tilde{V}^{l} &|& l\neq 1
\end{matrix}\right.
\end{equation}

\subsection{Transformer Temporal Extrapolator}

The core of Transformer lies in replacing the recursive structure with a multi-headed self-attentive mechanism.
The Transformer also follows the encoder-decoder structure widely used in RNNs.
The ability of Transformer to capture the nonlinear features of sequence data originates from the self-attention module, which allows Transformer to perform better time-dependent modeling relative to RNNs since RNNs only combine the current embedding with previously processed embeddings.
The self-attention mechanism first computes three vectors independently from the embeddings by linear mapping, which are query vectors ($\mathbf{Q}$), key vectors ($\mathbf{K}$), and value vectors ($\mathbf{V}$).
Then, the attention scores between embeddings are obtained by the scaled dot product of the query vector ($\mathbf{Q}$) and the key vector ($\mathbf{K}$). Finally, the attention is obtained from the normalized score and the value vector ($\mathbf{V}$) by multiplication and weighted summation.
\begin{equation}
\mathbf{Q}=f_{Q}(\mathbf{E}),~\mathbf{K}=f_{K}(\mathbf{E}),~\mathbf{V}=f_{V}(\mathbf{E})
\end{equation}
\begin{equation}
Attention(\mathbf{Q},\mathbf{K},\mathbf{V})=softmax(\frac{\mathbf{Q}\cdot \mathbf{K}^T}{\sqrt{d_k}})\mathbf{V}
\end{equation}

The superscript $T$ denotes the transpose of the vector, $d_k$ denotes the dimension of each query, and $\frac{1}{\sqrt{d_k}}$ is used to scale the dot product in order to stabilize the gradient.
The goal of the encoder is to encode the observation sequence into memory for the decoder's prediction; hence the key vector and value vector are output to the decoder after encoding the embedding in $1 \sim T_{obs}$ time.
The decoder regresses the future position distribution using the decoder's query vector and the encoder's output key vector and value vector for each time step in the $T_{obs}+1 \sim T_{obs}+T_{pred}$ prediction time.
The Transformer can learn long-time dependencies by capturing self-attention at each time step, and it is more suitable for prediction tasks than RNNs that record history with a limited single vector.
Moreover, Transformer decouples self-attention into three vectors of query, key, and value, contributing to learning more complex temporal dependencies.

The input embedding ($\mathbf{E}=V+P$) of the Transformer includes both the output of the previous Social Interaction Spatio-temporal Graph Convolutional Network ($V$) and also the positional encoding ($P$).
Positional encoding in the current prediction task refers to encoding timestamps. Each social interaction feature has a timestamp, and we form our input embedding by adding a feature that encodes a timestamp to the social interaction feature~\cite{Transformer_for_Trajectory}.
In the positional encoding calculation, each timestamp varies with the sine or cosine function according to the dimensions of the feature, which ensures that the entire sequence has a unique timestamp encoding.
As for positional encoding, Transformer has a significant difference from RNNs, where RNNs are input sequentially in temporal order, and the embedding position determines the order of processing; Transformer is parallel in processing and thus relies on positional encoding to represent the timestamp of the embedding.
Positional coding has an additional unique benefit of handling missing data for our prediction task.
The trajectory data are prone to missing position information sometimes, resulting in incomplete observation sequence data.
While the usual models ignore directly or complement by preprocessing through interpolation, Transformer-based models handle these missing data by automatically obtaining timestamp encoding information through positional encoding.
\begin{equation}
P^t={\begin{Bmatrix}
sin(\frac{t}{10000^{{d}/{D}}})~&|&~ d\in even\\ 
cos(\frac{t}{10000^{{d}/{D}}})~&|&~ d\in odd
\end{Bmatrix}}_{d=1}^D
\end{equation}

The multi-head attention mechanism can further improve the performance of the self-attention mechanism.
The multi-headed attention mechanism introduces multiple representation subspaces of self-attention and can jointly process these various subspaces.
It allows the model to focus on information from different positions by combining several different assumptions in calculating attention.
We first repeat $H$ times to extract attention, then connect them, and finally obtain multi-headed attention through a fully connected network $f$.

\begin{equation}
MultiHead(\mathbf{Q},\mathbf{K},\mathbf{V})=f(concat(head_1,\cdots,head_H))
\end{equation}
\begin{equation}
head_h=Attention_h(\mathbf{Q},\mathbf{K},\mathbf{V})
\end{equation}

\subsection{Multi-Modal Prediction}
Multi-modal prediction aims to generate probability distributions for the target agent's trajectories, which plays a crucial role in dealing with the uncertainty in motion trajectories.
Due to future uncertainty and multi-trajectory plausibility, the same agent can behave differently in the same scenario.
However, in realistic scenarios, there is only one ground truth trajectory data in each scene. So multi-modal prediction requires predicting the probability distribution of the target trajectory to cover all plausible trajectories.
Here, we follow the scheme proposed in~\cite{Social_LSTM} and assume that the future positions of agents obey a bivariate Gaussian distribution. The bivariate Gaussian distribution contains five parameters: the mean $\mu$, standard deviation $\sigma$, and correlation coefficient $\rho$. The trajectory of the $n$-th agent at time $t$ of the prediction is represented as follows.
\begin{equation}
    \hat{p}_{t}=(\hat{x}_{t},\hat{y}_{t}) \sim \mathcal{N}(\hat{\mu}_{t},\hat{\sigma}_{t},\hat{\rho }_{t})
\end{equation}
\begin{equation}
    \hat{\mu}_{t}=(\hat{\mu_{x}},\hat{\mu_{y}})_{t}
\end{equation}
\begin{equation}
    \hat{\sigma}_{t}=(\hat{\sigma_{x}},\hat{\sigma_{y}})_{t}
\end{equation}

The model is trained based on the ground truth trajectory and the predicted trajectory with Gaussian distribution, where the model parameters $\mathbf{W}$ are learned via minimizing the negative log-Likelihood loss.

{
\begin{equation}
    L_{pred}(\mathbf{W})=-\sum_{t=T_{obs}}^{T_{obs}+T_{pred}}log(\mathbb{}(p_{t}~|~\hat{\mu}_{t},\hat{\sigma}_{t},\hat{\rho }_{t}))
\end{equation}
}

{
Furthermore, we will analyze the collision rate in the trajectory prediction results to showcase our model's performance. Given that the training data predominantly comprises positive examples occurring in safe environments, it becomes challenging for the model to learn negative instances of collisions. Consequently, we adhere to the Social NCE~\cite{nce} benchmark and incorporate the Social NCE Loss into our model. This loss function is designed to minimize the likelihood of collisions within a trajectory by leveraging the learning of positive examples versus negative examples.
}

{
\begin{equation}
L_{nce}=-log \frac{ exp(emb_{hist} \cdot emb_{sample}^+ /\tau)  }
{ \sum_{sample}\sum_{n=0}^{N} exp(emb_{hist}\cdot emb_{sample}^n /\tau)}
\end{equation}
\begin{equation}
Loss=L_{pred}+\lambda L_{nce}
\end{equation}
}

{
In this context, $emb_{hist}$ serves as the query, representing the embedding of historical observations, while $emb_{sample}$ functions as the key, representing the sampling of future time points. In practice, we employ interpolation to perform 4 samplings per frame. This approach enables us to create one positive key along with multiple negative keys within the future trajectory. The loss function is then constructed by combining these positive and negative pairs involving the query and the keys.
And, $\tau$ represents the temperature factor, and $\lambda$ is the scaling factor. We utilize the benchmark settings of Social NCE to ensure a fair comparison.
}

\subsection{Algorithm}
The overall prediction process of our model is shown in Algorithm~\ref{algo}.
\begin{algorithm}[ht] 
\caption{{Attention-aware Social Graph Transformer Networks}}
\label{algo}
\begin{algorithmic}[1]
    \Require  
        The observed trajectory: $Traj_{obs}= \{p_{t}^{n}=(x_{t}^{n},y_{t}^{n})\}$.
    \Ensure  
        The probability distribution of the predicted trajectory:  $    \hat{p}_{t}^{n}=(\hat{x}_{t}^{n},\hat{y}_{t}^{n}) \sim \mathcal{N}(\hat{\mu}_{t}^{n},\hat{\sigma}_{t}^{n},\hat{\rho }_{t}^{n})$.
    
    \For{each $t\in [1,T_{obs}]$}  
      \State $G_{t} = (V_{t}, E_{t}) \gets Traj_{obs}$;
      \State ${G_{t}}' = ({V_{t}}', {A_{t}}') \gets G_{t}$;
    \EndFor  
    
    \State $\tilde{G}=\{ \tilde{V} ,\tilde{A} \}~|~
    \tilde{V}\in \mathbb{R}^{T\times K \times 2},~
    \tilde{A}\in \mathbb{R}^{T\times K \times K}
    \gets {G}'$;

    \If{Attention-aware}
        \State $att\text{-}\tilde{A} \gets (\tilde{A})$;
        \State $sp\text{-}att\text{-}\tilde{A} \gets att\text{-}\tilde{A}$;
    \EndIf
    
    \For{each $l \in [1,L]$}  
      \State $\tilde{V}^{l+1}=Conv_{temp}(Conv_{spa}(\tilde{G}^{l}))$;
    \EndFor  
    \State $\hat{\mu}_{t}^{n},\hat{\sigma}_{t}^{n},\hat{\rho }_{t}^{n}, emb_{hist}, emb_{sample}=Transformer(\tilde{V})$;
    \State \Return $\hat{\mu}_{t}^{n},\hat{\sigma}_{t}^{n},\hat{\rho }_{t}^{n}, emb_{hist}, emb_{sample}$;  
\end{algorithmic}
\end{algorithm}

\section{Experiments}

\subsection{Pedestrian Trajectory Prediction}

\subsubsection{Datasets and Evaluation Metrics}
We have adopted the same benchmark settings as~\cite{Social_LSTM, Social_GAN, nce}, aligning with the methodologies employed in most previous studies.
The benchmark datasets utilized in our study are ETH~\cite{ETH} and UCY~\cite{UCY}, encompassing videos capturing pedestrian trajectories in real-world scenarios characterized by high social interactions. These videos encompass four distinct scenarios and five datasets: ETH, HOTEL, UNVI, ZARA1, and ZARA2. Notably, ZARA1 and ZARA2 were recorded at different times but from the same location.
In terms of data sampling, the trajectory video data is collected at intervals of 0.4 seconds. The primary task involves observing 8 frames (equivalent to 3.2 seconds) and subsequently predicting the ensuing 12 frames (equivalent to 4.8 seconds). The real-world coordinates of each pedestrian can be derived from the pixel coordinates of the samples using the transformation matrix provided by the dataset.
To facilitate our evaluation, we have divided the study into five sub-tasks, each corresponding to one of the five sub-datasets. For each sub-dataset, the current dataset is designated as the test set, while the remaining datasets serve as the training and validation sets.
{
Our evaluation metrics encompass the Average Displacement Error (ADE), the Final Displacement Error (FDE) and Collision Rate (COL).
}

\begin{itemize}
\item Average Displacement Error (ADE): Also known as Mean Squared Error (MSE), this metric calculates the average displacement error between the predicted trajectory and the ground truth trajectory.
\item Final Displacement Error (FDE): This metric quantifies the displacement error between the predicted position and the ground truth position in the last frame.
\item {Collision Rate (COL): The percentage of test cases in which the predicted trajectories of different agents result in collisions.}
\end{itemize}


The ultimate output of our model is the distribution representing the predicted trajectory. In line with established benchmark methods, we calculate the Average Displacement Error (ADE) and Final Displacement Error (FDE) by generating 20 samples from the distribution and subsequently identifying the position closest to the ground truth to compute the displacement error.

\subsubsection{Quantitative Analysis}

We perform experiments on five distinct subtasks separately and compare the results against baselines, as detailed in Table~\ref{table_result}.
For consistency and fairness in our evaluation, all experiments are conducted following a leave-one-out cross-validation strategy established by the benchmark.

{
We compared the results of our model with Linear, LSTM, Social LSTM~\cite{Social_LSTM}, Social ATTN~\cite{Social_Att}, Social GAN~\cite{Social_GAN}, SoPhie~\cite{SoPhie}, Social-BiGAT~\cite{Social-BiGAT}, Social-STGCNN~\cite{Social-STGCNN}, Transformer~\cite{Transformer_for_Trajectory}, STAR~\cite{STAR}, Trajectron~\cite{Trajectron}, Trajectron++\cite{Trajectron++}, DMRGCN~\cite{DMRGCN}, generativeSCAN~\cite{generativeSCAN} and Social NCE~\cite{nce}. Descriptions of the models are given in Appendix 1.
SGTN serves as our baseline for Social Graph Transformer Networks, A-SGTN is the variant that incorporates the attention mechanism, and SA-SGTN is the variant that includes the sparse attention mechanism.
}

\begin{table}[ht]
\caption{Pedestrian Trajectory Prediction Results. Values denote ADE/FDE in meters; * denotes non-multi-modal prediction model; ${}^\dagger$ denotes additional information such as images. The best results are indicated in bold, while the second-best results are underlined.}
\label{table_result}
\centering
\resizebox{0.9\linewidth}{!}{
\begin{tabular}{l|cccccc}
\toprule
                                                & ETH        & HOTEL     & UNIV      & ZARA1     & ZARA2     & AVG       \\ \midrule
Linear*                                         & 1.33/2.94  & 0.39/0.72 & 0.82/1.59 & 0.62/1.21 & 0.77/1.48 & 0.79/1.59 \\ \midrule
LSTM*                                           & 1.09/2.41  & 0.86/1.91 & 0.61/1.31 & 0.41/0.88 & 0.52/1.11 & 0.70/1.52 \\ \midrule
Social LSTM*~\cite{Social_LSTM}                 & 1.09/2.35  & 0.79/1.76 & 0.67/1.40 & 0.47/1.00 & 0.56/1.17 & 0.72/1.54 \\ \midrule
Social ATTN*~\cite{Social_Att}                  & \underline{0.39}/3.74  & 0.29/2.64 & 0.33/3.92 & 0.20/0.52 & 0.30/2.13 & 0.30/2.59 \\ \midrule
Social GAN~\cite{Social_GAN}                    & 0.81/1.52  & 0.72/1.61 & 0.60/1.26 & 0.34/0.69 & 0.42/0.84 & 0.58/1.18 \\ \midrule
SoPhie${}^\dagger$~\cite{SoPhie}                & 0.70/1.43  & 0.76/1.67 & 0.54/1.24 & 0.30/0.63 & 0.38/0.78 & 0.54/1.15 \\ \midrule
Social-BiGAT~\cite{Social-BiGAT}                & 0.69/1.29  & 0.49/1.01 & 0.55/1.32 & 0.30/0.62 & 0.36/0.75 & 0.48/1.00 \\ \midrule
Social-STGCNN~\cite{Social-STGCNN}              & 0.64/1.11  & 0.49/0.85 & 0.44/0.79 & 0.34/0.53 & 0.30/0.48 & 0.44/0.75 \\ \midrule
Transformer~\cite{Transformer_for_Trajectory}   & 0.61/1.12  & 0.18/0.30 & 0.35/0.65 & 0.22/0.38 & 0.17/0.32 & 0.31/0.55 \\ \midrule
STAR~\cite{STAR}                                & \textbf{0.36}/0.65  & 0.17/0.36 & 0.26/0.55 & 0.22/0.46 & 0.31/0.62 & 0.26/0.53 \\ \midrule
Trajectron~\cite{Trajectron}                    & 0.59/1.14  & 0.35/0.66 & 0.54/1.13 & 0.43/0.83 & 0.43/0.85 & 0.47/0.92 \\ \midrule
Trajectron++~\cite{Trajectron++}                & \underline{0.39}/0.83  & \textbf{0.12}/0.21 & 0.20/0.44 & \textbf{0.15}/0.33 & \textbf{0.11}/0.25 & \textbf{0.19}/0.41 \\ \midrule
DMRGCN~\cite{DMRGCN}                            & 0.60/1.09 & 0.21/0.30 & 0.35/0.63 & 0.29/0.47 & 0.25/0.41 & 0.34/0.58 \\ \midrule
generativeSCAN~\cite{generativeSCAN}            & 0.79/1.49 & 0.37/0.74 & 0.58/1.23 & 0.37/0.78 & 0.31/0.66 & 0.48/0.98 \\ \midrule \midrule
Ours (SGTN)                                     & 0.40/\underline{0.59}  & \underline{0.14}/\textbf{0.16} & \underline{0.16}/\textbf{0.18} & \underline{0.17}/0.29 & \underline{0.14}/\underline{0.17} & \underline{0.20}/\underline{0.28} \\ \midrule
{Ours (A-SGTN)  }                                 & 0.41/\textbf{0.54}  & 0.18/0.24 & \underline{0.16}/\textbf{0.18} & \underline{0.17}/\textbf{0.22} & 0.15/0.19 & 0.21/\textbf{0.27} \\ \midrule
{Ours (SA-SGTN)}                                  & 0.48/0.62  & 0.18/\underline{0.20} & \textbf{0.15}/\underline{0.19} & 0.19/\underline{0.28} & \underline{0.14}/\textbf{0.16} & 0.23/0.29 \\ \bottomrule

\end{tabular}
}
\end{table}

\begin{table}[ht]
\caption{{Trajectory prediction collision rate results. Values are percentages of collision rate. The best results are bolded.}}
\centering
\label{table_nce}
\resizebox{0.9\linewidth}{!}{
\begin{tabular}{l|cccccc}
 \toprule
COL                                         & ETH  & HOTEL & UNIV & ZARA1 & ZARA2 \\ \midrule
Social-STGCNN ~\cite{Social-STGCNN}         & 1.33 & 3.82  & 9.11 & 2.27  & 6.86  \\ \midrule
Social-STGCNN w/ nce~\cite{nce}             & 0.61 & 3.35  & 6.44 & 1.02  & 3.37  \\ \midrule
Trajectron++ ~\cite{Trajectron++}           & 1.16 & 0.84  & 3.38 & 0.46  & 1.03  \\ \midrule
Trajectron++ w/ nce~\cite{nce}              & \textbf{0.00} & 0.38  & 3.08 & 0.18  & \textbf{0.99}  \\ \midrule
ours (SGTN)                                 & \textbf{0.00} & 0.34  & 4.89 & 0.44  & 1.94  \\ \midrule
ours (SGTN) w/ nce                          & \textbf{0.00} & \textbf{0.28}  & \textbf{2.96} & \textbf{0.13}  & 1.14  \\ \bottomrule
\end{tabular}
}
\end{table}

Overall, our model achieves the state-of-the-art results in FDE across all datasets, holds the state-of-the-art position in ADE for the UNIV dataset, and delivers comparable ADE results for the other four datasets.
Notably, we observe that the simple linear model outperforms some deep learning models on the HOTEL dataset. 
This can be attributed to HOTEL predominantly containing linear trajectory data. In contrast to other datasets, HOTEL exhibits lower pedestrian density, making it more suitable for simple linear models. Complex deep learning models tend to overfit and experience performance degradation in such cases.
The Trajectron++ model excels in optimizing the three datasets based on the ADE measure. However, our model achieves a substantial decrease in FDE while maintaining a close ADE. 
This demonstrates that our Transformer-based model outperforms the LSTM-based Trajectory++ model in long-term prediction. 
While the LSTM-based model makes accurate predictions in the immediate future, its displacement error gradually increases as the time step grows due to its limitations in handling long-term historical memory. In contrast, our model consistently maintains a small error in long-term predictions.
Our model attains the optimal ADE with FDE on the UNVI dataset. This observation highlights that UNVI, being the most densely populated dataset, benefits significantly from our Social Interaction Spatio-temporal Graph, enhancing our model's ability to gather information in complex scenarios.
{
Scenarios for each dataset are shown in Appendix 1.
}
In conclusion, the average ADE and FDE metrics indicate that our baseline model reduces FDE by 0.13 meters compared to Trajectron++, a decrease of nearly 31.7\%, while maintaining an average ADE only slightly higher than the state-of-the-art by 0.01 meters.

{
Furthermore, we present a comparison of our baseline model with two variants: A-SGTN and SA-SGTN.
A-SGTN demonstrates further optimization in the FDE metrics across the three datasets, courtesy of the attention mechanism's assistance. This suggests that the attention mechanism can intelligently collect social interaction information, thereby reducing the final displacement error.
In the final average results, we observe that A-SGTN has a slightly higher ADE compared to SGTN, while the FDE is slightly lower than that of SGTN.
On the other hand, SA-SGTN effectively prunes low-weighted social interaction information through sparse attention processing. This results in reduced ADE in the UNIV dataset and FDE in ZARA2 dataset. This indicates that the judicious application of sparse attention aids in reducing redundant social interaction information and, consequently, enhances prediction accuracy.
However, it's worth noting that for datasets with inherently lower pedestrian density, sparse attention may inadvertently remove crucial social interaction information, leading to less favorable outcomes.
}

{
We followed the settings of Social NCE~\cite{nce} and compared the collision rates of pedestrian trajectory prediction results when the distance hyperparameter was set to 0.2 meters. We provide experimental results for our baseline model (SGTN) with and without the addition of Social Contrastive Loss (NCE Loss), as shown in Table~\ref{table_nce}. Notably, our SGTN achieves a test result of 0 collisions in the ETH dataset without any changes, a result that outperforms both the original Social-STGCNN and Trajectron++. Even when Social-STGCNN with NCE Loss is added, our SGTN still outperforms it, while Trajectron++ with NCE Loss is comparable. This indicates that our SGTN prediction results are accurate enough to reduce the occurrence of collisions. Additionally, on the HOTEL dataset, it outperforms all other benchmarks without changes. Our model with the addition of NCE Loss improves on all collision rate reductions and achieves optimal results on four datasets.
}

\subsubsection{Qualitative Analysis}

\begin{figure}[th]
  \centering
  \includegraphics[width=\linewidth]{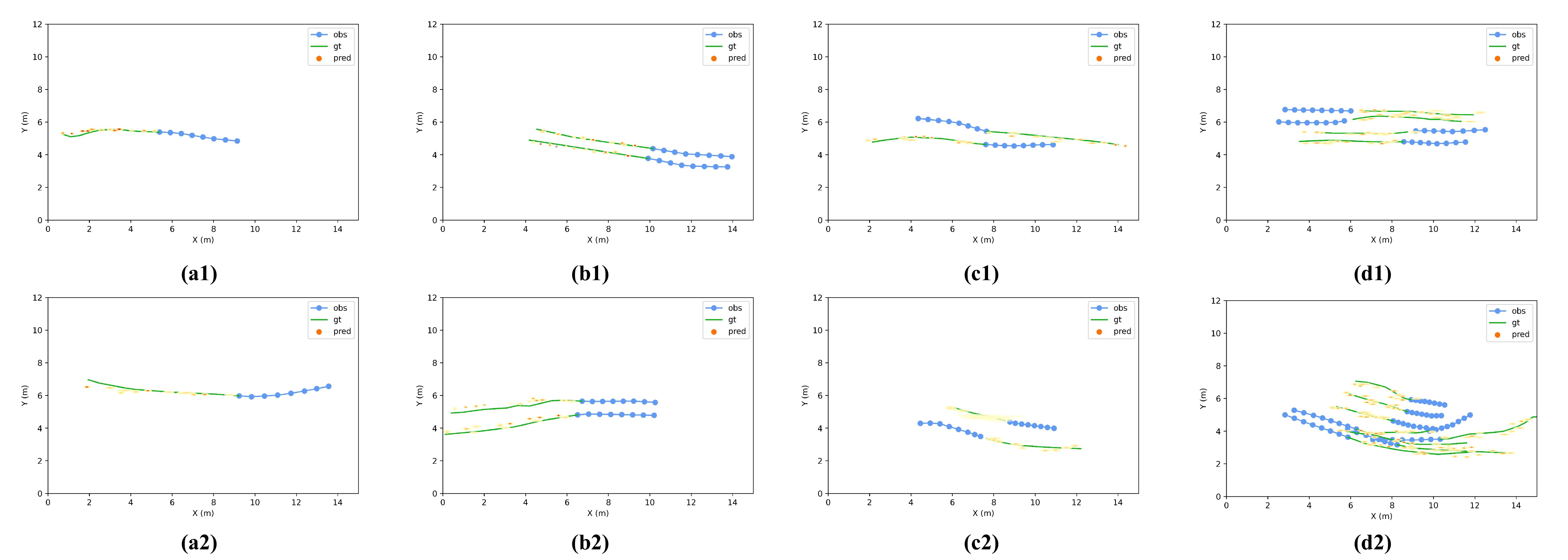}
  \caption{Multi-modal Prediction of Pedestrian Trajectories. (a1), (a2) for single person; (b1), (b2) for walking in the same direction; (c1), (c2) for walking in opposite directions; (d1), (d2) for mixed multi-person. The predicted positions are probability distributions, with darker colors indicating higher probabilities.}
  \label{figure-gaussian}
\end{figure}

\begin{figure}[th]
  \centering
  \includegraphics[width=\linewidth]{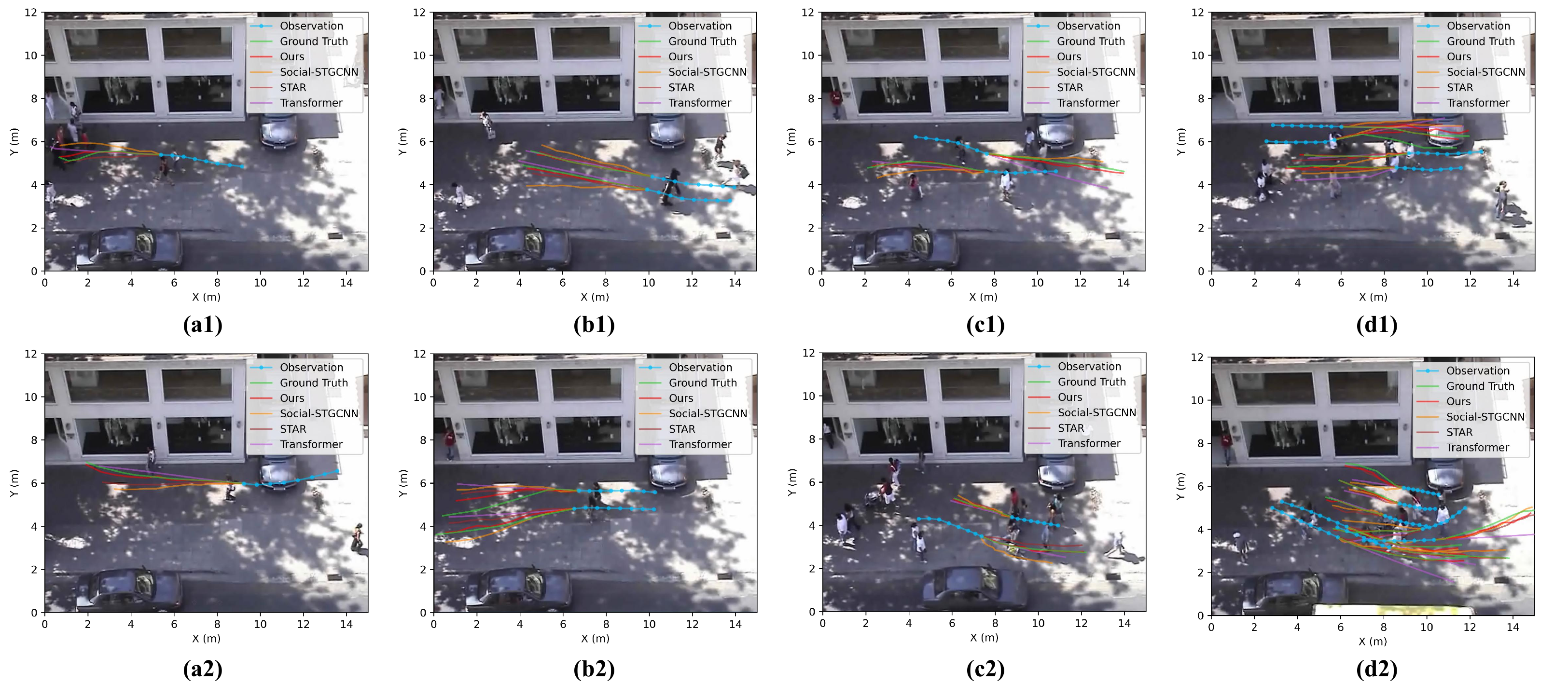}
  \caption{Trajectory Prediction Visualization Comparison. (a1), (a2) for single person; (b1), (b2) for walking in the same direction; (c1), (c2) for walking in opposite directions; (d1), (d2) for mixed multi-person.}
  \label{figure-comparison}
\end{figure}

Figure~\ref{figure-gaussian} illustrates the results of multi-modal pedestrian trajectory predictions for 4 different scenarios.
In the figure, the observed trajectories are represented by blue lines, while the ground truth trajectories are shown in green. For each time step of the prediction, we generate multi-modal predictions, with darker colors indicating higher probabilities.
(a1) and (a2) depict scenarios where a single pedestrian walks independently. In these cases, our predicted trajectories closely align with the ground truth trajectories, exhibiting similar bending tendencies.
(b1) and (b2) showcase scenes involving two individuals walking in the same direction. Here, the two pedestrians maintain consistent direction, speed, and nearly fixed distance. Our prediction results effectively capture this scenario.
(c1) and (c2) portray scenarios featuring two pedestrians walking in opposite directions, often involving collision avoidance maneuvers. The predicted probability distribution from our model accurately reflects the awareness of collision avoidance.
(d1) and (d2) depict mixed pedestrian scenarios, where trajectories demonstrate that individuals within the same group tend to approach closely, while those from different groups maintain distance to avoid collisions. Our model's probability distribution effectively captures these dynamics.
These four scenarios collectively demonstrate our model's ability to handle pedestrian trajectory prediction across a spectrum of scenarios, ranging from simple to complex.

Figure~\ref{figure-comparison}, corresponding to Figure~\ref{figure-gaussian}, offers a visual comparison of prediction results between our model and three other advanced models.
In summary, the Transformer model tends to produce approximately straight-line trajectories in each scenario, as it does not consider social interaction information. It excels in near-linear trajectory scenarios but struggles to predict accompanying walks and collision avoidance.
The Social-STGCNN model relies on graph structures and convolutional networks for predictions, avoiding the use of recurrent networks and Transformers. However, its prediction errors are larger compared to STAR and our model. 
In several scenarios, such as (c1) and (c2), Social-STGCNN exhibits significant errors, especially in the last few frames, due to its approach that doesn't rely on sequential data. Meanwhile, the Social-STGCNN predicts undeserved avoidance for peers in (b1) scenario and over-avoidance when avoiding collisions in (c2) scenario.
Moreover, it tends to maintain a large distance in multi-person scenes, lacking the ability for long-term predictions.
Both STAR and our model leverage graph structures and Transformers, offering substantial advantages over the other two models.
Compared to STAR, our model not only employs graphs for data organization but also uses the adjacency matrix to gather social interaction information and generate pseudo-images, enhancing Transformer performance. This provides an edge in various scenarios, particularly complex ones like (d1) and (d2), where our predictions closely align with the ground truth.
See Appendix 1 for more visualizations.
In summary, our model excels in collecting social interaction information through the adjacency matrix, comprehending scenarios involving peers and collision avoidance, and optimizing Transformer performance with pseudo-images, making it particularly well-suited for long-term predictions.

\subsubsection{Ablation Experiments}

To optimize our model's hyperparameters, we conducted ablation experiments on the ETH dataset. Table~\ref{table_struc} presents the results of these experiments, which focused on the model structure.
Our model employs the Social Spatio-temporal Graph Convolutional Network (SSTGCN), where both the inputs and outputs remain constant. This allows for the use of multiple layers for feature extraction. Surprisingly, we found that a single layer of SSTGCN yielded the best results. This can be attributed to the fact that repeated Spatio-temporal feature extraction can introduce redundancy, potentially diminishing prediction accuracy.
Additionally, we explored various hyperparameters for the Transformer. The optimal results were achieved when utilizing a multi-head mechanism with 4 heads, a stack comprising 6 layers of encoding and decoding modules, an embedding feature dimension of 8, and a feedforward network dimension of 32.
It's important to note that our chosen hyperparameter settings for the Transformer are relatively modest in comparison to a standard Transformer network. As demonstrated by other models in Table~\ref{table_result}, trajectory prediction feature extraction typically doesn't require extremely high dimensions. Overly complex models can lead to overfitting, which in turn increases prediction errors.

{
We subsequently conducted ablation experiments on SSTGCN to assess spatial and temporal information, the results of which are presented in Table~\ref{table_graph}. 
For the spatial information, we evaluated the impact of the adjacency matrix. Using this matrix allows us to aggregate social interaction information. In its absence, individual trajectory features are not aggregated and are directly forwarded to the subsequent phase.
In terms of temporal information, this was gauged by manipulating the convolutional kernel to determine if sequence features should be aggregated. Cases without temporal information only encompass the current frame's data, while those with temporal information, by default, incorporate data from every frame preceding and succeeding the current one.
The outcomes reveal that employing SSTGCN with integrated spatio-temporal feature aggregation yields the most accurate predictions. This underscores the efficacy of our SSTGCN design in enhancing trajectory prediction accuracy.
}

\begin{table}[ht]
\caption{Ablation experiments for model structure on ETH. SSTGCN denotes the number of layers; Others denote the hyperparameters of Transformer. Values denote ADE/FDE in meters. The best results are indicated in bold.}
\centering
\label{table_struc}
\resizebox{0.9\linewidth}{!}{
\begin{tabular}{c|cccc|c}
\toprule
SSTGCN & Multi-Head & Layers & Embedding  & Feedforward   & ADE/FDE     \\ \midrule
1      & 4          & 4      & 8        & 32        & 0.53/0.79     \\ \midrule
1      & 4          & 4      & 16       & 64        & 0.47/0.61     \\ \midrule
1      & 4          & 6      & 8        & 32        & \textbf{0.40/0.59}     \\ \midrule
1      & 4          & 6      & 16       & 64        & 0.47/0.66     \\ \midrule
1      & 4          & 6      & 32       & 128       & 0.73/1.27     \\ \midrule
1      & 8          & 6      & 16       & 64        & 0.53/0.87     \\ \midrule
1      & 8          & 6      & 32       & 128       & 0.59/1.04     \\ \midrule
1      & 8          & 6      & 64       & 256       & 0.82/1.43     \\ \midrule
3      & 4          & 6      & 8        & 32        & 0.55/0.83     \\ \midrule
3      & 4          & 6      & 16       & 64        & 0.58/0.98     \\ \midrule
3      & 4          & 6      & 32       & 128       & 0.61/0.94      \\ \midrule
3      & 8          & 6      & 32       & 128       & 0.99/1.48      \\ \midrule
5      & 4          & 6      & 16       & 64        & 0.56/0.90      \\ \midrule
5      & 4          & 6      & 32       & 128       & 0.63/1.01       \\ \bottomrule
\end{tabular}
}
\end{table}

\begin{table}[ht]
\caption{Ablation experiments for the SSTGCN on ETH. Values denote ADE/FDE in meters. The best results are indicated in bold. }
\centering
\label{table_graph}
\resizebox{0.8\linewidth}{!}{
\begin{tabular}{cc|cc}
\toprule
  Spatial Features   & Temporal Features     & ADE    & FDE    \\
\midrule
  w/o & w/o & 0.64 & 0.99 \\
\midrule
  w/  & w/o & 0.54 & 0.81 \\
\midrule
  w/o & w/ & 0.46 & 0.71 \\
\midrule
  w/  & w/ & \textbf{0.40} & \textbf{0.59} \\
\bottomrule
\end{tabular}
}
\end{table}

\subsection{Vehicle Trajectory Prediction}

\subsubsection{Datasets and Evaluation Metrics}
We validate the results of vehicle trajectory prediction on publicly available datasets NGSIM I-80~\cite{80} and US-101~\cite{101}. These two datasets are real vehicle trajectory datasets for two U.S. highways, each containing three sub-datasets for light, moderate, and congested traffic conditions. Each sub-dataset has a duration of 15 minutes and is sampled at 10 Hz, containing the local coordinates of all vehicles.
We follow the benchmark practice of downsampling the dataset by a factor of 2 and then slicing it into 8-second segments, with 3 seconds (15 frames) for observation and 5 seconds (25 frames) for prediction. A quarter of the data in each sub-dataset is used for testing, and the rest is used for training.
We evaluate the model performance using the root mean square error (RMSE), with the evaluation metric covering RMSE from the 1st to the 5th second.

\begin{table}[ht]
\caption{Vehicle Trajectory Prediction Results. Values denote RMSE in meters. GRIP and our model predict the trajectory of all vehicles simultaneously, while other models predict the trajectory of a given vehicle (middle position).}
\centering
\label{table_vehicle_result}
\resizebox{\linewidth}{!}{
\begin{tabular}{c|cccccc|c|ccc}
\toprule
\begin{tabular}[c]{@{}c@{}}Prediction\\ Horizon(s)\end{tabular}   & CV   & V-LSTM & \begin{tabular}[c]{@{}c@{}}C-CGMM\\ +VIM~\cite{cslstm-pre}\end{tabular}   & \begin{tabular}[c]{@{}c@{}}GAIL\\ -GRU~\cite{gru-compare}\end{tabular}  & \begin{tabular}[c]{@{}c@{}}CS\\ -LSTM~\cite{cslstm}\end{tabular} & \begin{tabular}[c]{@{}c@{}}CS\\ -LSTM(M)~\cite{cslstm}\end{tabular} & GRIP~\cite{GRIP} & SGTN & A-SGTN & {SA-SGTN} \\ \midrule
1                      & 0.73 & 0.68   & 0.66       & 0.69     & 0.61    & 0.62       & 0.64 & 0.59  & 0.55 & \textbf{0.52} \\ \midrule
2                      & 1.78 & 1.65   & 1.56       & 1.51     & 1.27    & 1.29       & 1.13 & 1.13  & 1.01 & \textbf{0.92} \\ \midrule
3                      & 3.13 & 2.91   & 2.75       & 2.55     & 2.09    & 2.13       & 1.80 & 1.61  & 1.43 & \textbf{1.23} \\ \midrule
4                      & 4.78 & 4.46   & 4.24       & 3.65     & 3.10    & 3.20       & 2.62 & 2.04  & 1.84 & \textbf{1.61} \\ \midrule
5                      & 6.68 & 6.27   & 5.99       & 4.71     & 4.37    & 4.52       & 3.60 & 2.45  & 2.21 & \textbf{1.76} \\ \bottomrule
\end{tabular}
}
\end{table}

\begin{figure}[th]
  \centering
  \includegraphics[width=0.95\linewidth]{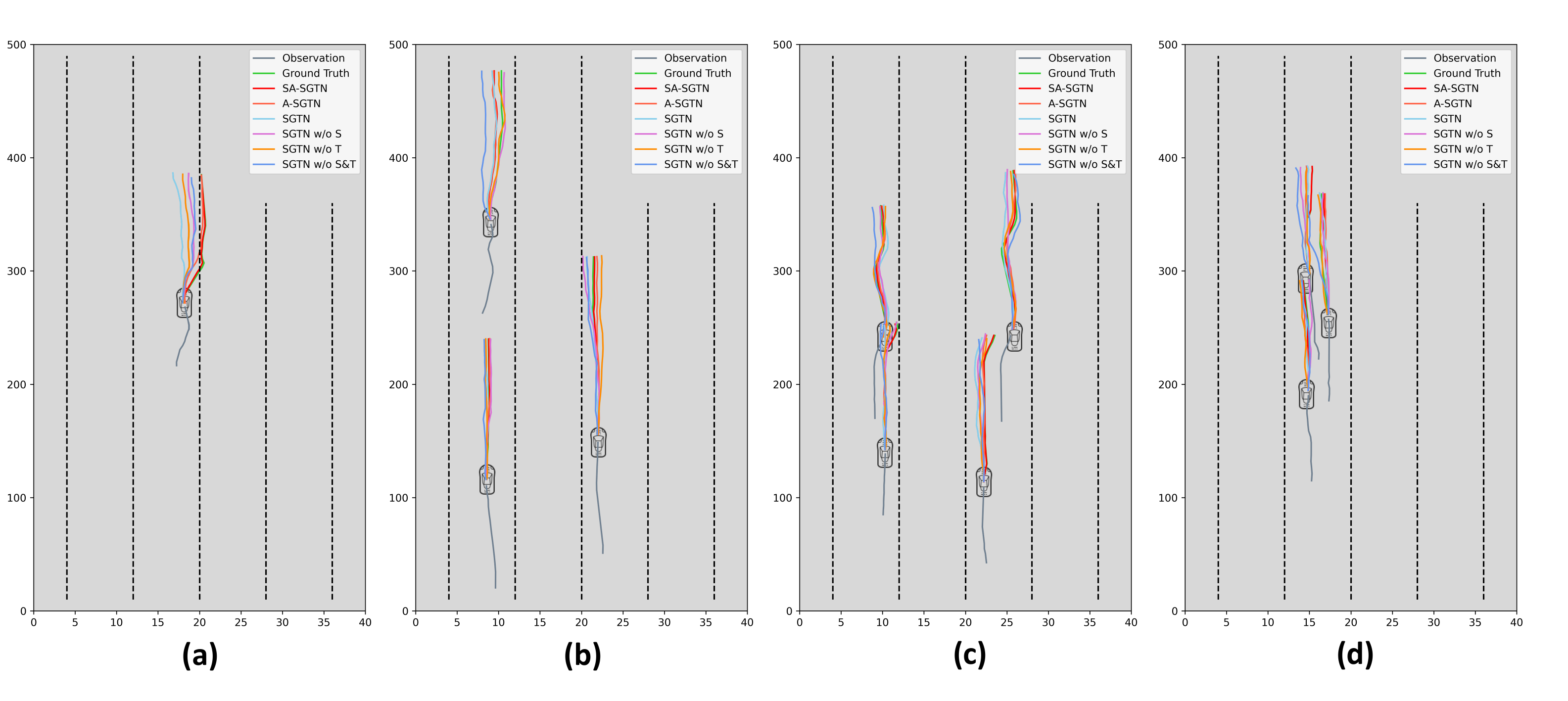}
  \caption{{Ablation visualization. It contains predicted vehicle trajectories for different models in four scenarios.}}
  \label{figure-road}
\end{figure}

\subsubsection{Comparison Experiments}

{
We compare our predicted results with the following baselines:CV (Constant Velocity), V-LSTM (Vanilla LSTM), C-VGMM+VIM~\cite{cslstm-pre}, GAIL-GRU~\cite{gru-compare}, CS-LSTM~\cite{cslstm} and GRIP~\cite{GRIP}. Descriptions of the models are given in Appendix 2.
Our model consists of the baseline Social Graph Transformation Network (SGTN) and two variants: Attention-Aware Social Graph Transformation Network (A-SGTN) and Sparse Attention-Aware Social Graph Transformation Network (SA-SGTN).
Our models, like GRIP, predict all vehicle trajectories simultaneously. 
Similar to GRIP, our models predict all vehicle trajectories simultaneously and are multi-modal prediction methods. To ensure a fair comparison, we perform 6 sampling in multi-modal prediction, referring to the 6 maneuvers used in CS-LSTM.
}

Table~\ref{table_vehicle_result} shows the comparison of our model's and other models' prediction results (obtained from~\cite{GRIP,cslstm}).
The Constant Velocity model, which serves as the simplest baseline, achieves an RMSE of 0.73 at the 1st second and 6.68 at the 5th second. Subsequent RNN-based LSTM and GRU methods gradually decrease RMSE but still have limitations.
These models predict the trajectories of middle vehicles by aggregating information from surrounding vehicles. However, they can only predict the trajectories of vehicles at one specified middle location at a time and cannot predict all vehicle trajectories simultaneously.
In contrast, GRIP significantly enhances accuracy and practicality by predicting all vehicle trajectories simultaneously, resulting in a notable reduction in RMSE.
Our SGTN improves accuracy while also predicting all vehicle trajectories, establishing itself as state-of-the-art across all metrics.
Furthermore, our A-SGTN model, incorporating the Attention-aware module, further boosts accuracy.
A-SGTN reduces RMSE by 0.04, 0.12, 0.18, 0.20, and 0.24 (respectively, 5 seconds) compared to SGTN, underscoring the contribution of the Attention-aware module to distal time prediction accuracy.
A-SGTN also reduces RMSE relative to GRIP by 14\%, 10.6\%, 20.6\%, 29.8\%, and 38.6\% (respectively, 5 seconds). 
This progressive improvement highlights the suitability of our model for long-term prediction scenarios.

{
Furthermore, we achieved enhanced prediction accuracy through the incorporation of sparse attention in our SA-SGTN. This highlights a unique advantage of the attention mechanism in vehicle trajectory prediction. Unlike pedestrian trajectory prediction, where pedestrians typically move at slower speeds and often need to be aware of their surroundings in complex social interactions, vehicle movement, particularly on highways, involves higher speeds, and drivers maintain focused attention on specific aspects of their environment.
For instance, when changing lanes, a driver focuses on vehicles ahead and behind in the same lane while disregarding those on the opposite side. During acceleration, a driver pays attention to vehicles in front and those merging from the left and right but may disregard those behind.
In conclusion, it can be observed that attention mechanisms, especially sparse attention mechanisms, have a more pronounced positive impact on vehicle trajectory prediction compared to pedestrian trajectory prediction. This suggests that vehicle trajectory features exhibit significant differences from those of pedestrians, underscoring the importance of considering attention mechanisms in vehicle trajectory prediction tasks.
}

\subsubsection{Ablation and Visualization}

The success of our vehicle trajectory prediction model is primarily attributed to the Social Interaction Spatio-Temporal Graph and the attention mechanisms. To provide a comprehensive comparison of our model's components, we conducted an ablation study and visualized the results.


{
In this study, we examined various model variants: `S' and `T' denote spatial and temporal features, respectively.
We carefully selected specific scenarios for visualization and comparison of the prediction results, as depicted in Figure~\ref{figure-road}.
See Appendix 2 for more visualizations.
These visualizations allow us to assess the performance of each model variant in various scenarios and provide valuable insights into the significance of spatial and temporal feature aggregation, as well as the impact of attention mechanisms.
}

{
In Figure~\ref{figure-road}(a)-(d), we compare four scenarios: solo vehicle, light, moderate, and congested traffic conditions, respectively.
Figure~\ref{figure-road}(a) illustrates a situation where the vehicle attempts to move to the right side. Notably, among all models, only A-SGTN and SA-SGTN come closer to predicting this trajectory accurately. In contrast, all other models exhibit predictions that are generally in the forward direction.
In Figure~\ref{figure-road}(b), three vehicles are widely spaced and move forward simultaneously. The SGTN w/o S\&T model, lacking Spatio-temporal feature collection, yields the poorest prediction results. Additionally, SGTN w/o S and SGTN w/o T models also deviate somewhat from the ground truth trajectory.
Figure~\ref{figure-road}(c) represents moderate congestion, with vehicles traveling at reduced speeds and displaying left and right zigzagging trajectories. When compared to the SGTN model, the results of A-SGTN and SA-SGCN align more closely with the ground truth trajectory.
Lastly, Figure~\ref{figure-road}(d) portrays severe congestion, with vehicles moving slowly forward and predominantly in a straight line. Models lacking Spatio-temporal features show increased deviations, especially the SGTN w/o S\&T models.
In summary, the SA-SGTN model aligns most closely with the ground truth trajectory, followed by the A-SGTN model.
}

\subsection{Hybrid Trajectory Prediction}

\subsubsection{Datasets and Evaluation Metrics}

{
We utilized the ApolloScape trajectory dataset~\cite{TrafficPredict} for our hybrid trajectory prediction task. The ApolloScape trajectory dataset is derived from peak-hour urban traffic data collected by the Apollo system. This data was gathered at traffic intersections and mixed pedestrian-vehicle roadways, encompassing various types of traffic agents such as vehicles, pedestrians, and bicyclist (including motorcyclists).
The dataset comprises 53 minutes of training data, providing agent locations, sizes, heading angles, and other relevant information. The data was sampled at a rate of 2 frames per second. The prediction task involved observing the past 6 frames and forecasting the subsequent 6 frames into the future.
While the official ApolloScape test set requires verification on their website, it is currently unavailable. Hence, we follow a common practice of allocating 20\% of the original training dataset as the test set, while the remaining data is split into a new train set and a validation set.
}

{
The evaluation metrics still employ the Average Displacement Error (ADE) and the Final Displacement Error (FDE), similar to those used in pedestrian trajectory prediction. Given the diverse types of agents in the ApolloScape trajectory dataset, in accordance with the configuration of the benchmark~\cite{TrafficPredict}, the final composite metrics are compared using a weighted summation, which includes the weighted sum of ADE (WSADE) and the weighted sum of FDE (WSFDE).
}
{
\begin{equation}
WSADE=0.20 \cdot ADE_v+0.58 \cdot ADE_p+0.22 \cdot ADE_b
\end{equation}
\begin{equation}
WSFDE=0.20 \cdot FDE_v+0.58 \cdot FDE_p+0.22 \cdot FDE_b
\end{equation}
In these equations, `v' represents vehicles, `p' represents pedestrians, and `b' represents bicyclists (including motorcyclists). 
}

\subsubsection{Comparison Experiments}

{
We compared our predictions with those of the following methods: TraﬃcPredict~\cite{TrafficPredict} , Social LSTM~\cite{Social_LSTM} , Social GAN~\cite{Social_GAN} , STAR~\cite{STAR} . Constant Velocity, StarNet~\cite{StarNet}, Transformer~\cite{Transformer_for_Trajectory}, TPNet~\cite{TPNet}, GRIP++~\cite{GRIP++} and S2TNet~\cite{S2TNet}.
Descriptions of the models are given in Appendix 3.
We provide predictions for the baseline SGTN as well as the two variants, A-SGTN and SA-SGTN. Since current benchmarks typically do not involve multi-modal predictions, we also adhere to the benchmark for direct predictions in this task.
}

\begin{table}[ht]
\caption{{Hybrid Trajectory Prediction Results. WSADE denotes weighted sum of ADE, WSFDE denotes weighted sum of FDE, and lowercase letters v, p, b denote vehicle, pedestrian, and bicyclist (motorcyclist), respectively. Values denote displacement error in meters. The best results are indicated in bold, while the second-best results are underlined.}}
\centering
\label{table_hybrid_result}
\resizebox{0.9\linewidth}{!}{
\begin{tabular}{l|c|ccc|c|ccc}
\toprule
Method                                          & WSADE & ADEv & ADEp & ADEb  & WSFDE & FDEv  & FDEp  & FDEb  \\ \midrule
TraﬃcPredict~\cite{TrafficPredict}              & 8.59  & 7.95 & 7.18 & 12.88 & 24.23 & 12.78 & 11.12 & 22.79 \\ \midrule
Social LSTM~\cite{Social_LSTM}                  & 1.90  & 2.95 & 1.29 & 2.53  & 3.40  & 5.28  & 2.32  & 4.54  \\ \midrule
Social GAN~\cite{Social_GAN}                    & 1.59  & 3.04 & 0.98 & 1.84  & 2.78  & 5.09  & 1.73  & 3.45  \\ \midrule
STAR~\cite{STAR}                                & 1.55  & 2.56 & 0.95 & 2.17  & 2.86  & 4.63  & 1.80  & 4.04  \\ \midrule
Constant Velocity                               & 1.48  & 2.65 & 0.85 & 2.05  & 2.76  & 4.79  & 1.64  & 3.86  \\ \midrule
StarNet~\cite{StarNet}                          & 1.34  & 2.39 & 0.79 & 1.86  & 2.50  & 4.29  & 1.52  & 3.46  \\ \midrule
Transformer~\cite{Transformer_for_Trajectory}   & 1.28  & 2.23 & 0.73 & 1.84  & 2.40  & 4.03  & 1.43  & 3.48  \\ \midrule
TPNet~\cite{TPNet}                              & 1.28  & 2.21 & 0.74 & 1.85  & \underline{2.34}  & 3.86  & 1.41  & 3.40  \\ \midrule
GRIP++~\cite{GRIP++}                            & 1.26  & 2.24 & \underline{0.71} & 1.80  & 2.36  & 4.08  & \underline{1.37}  & 3.42  \\ \midrule
S2TNet~\cite{S2TNet}                            & \underline{1.17}  & 1.99 & \textbf{0.68} & 1.70  & \textbf{2.18}  & 3.58  & \textbf{1.30}  & 3.21  \\ \midrule \midrule
Ours (SGTN)                                     & 1.19  & 1.35 & 0.91 & 1.18  & 2.52  & 2.75  & 1.98  & 2.51  \\ \midrule
Ours (A-SGTN)                                   & \textbf{1.14}  & \underline{1.28} & 0.85 & \textbf{1.14}  & 2.38  & \underline{2.62}  & 1.82  & \textbf{2.43}  \\ \midrule
Ours (SA-SGTN)                                  & \textbf{1.14}  & \textbf{1.25} & 0.90 & \underline{1.15}  & 2.40  & \textbf{2.59}  & 1.95  & \underline{2.46}  \\ \bottomrule
\end{tabular}
}
\end{table}

\begin{figure}[th]
  \centering
  \includegraphics[width=0.7\linewidth]{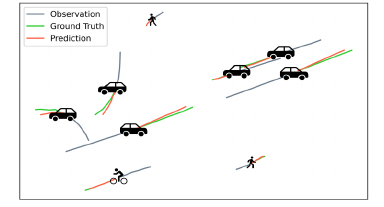}
  \caption{Visualization of Hybrid trajectory predictions.}
  \label{figure-mix}
\end{figure}

{
The results of hybrid trajectory prediction are presented in Table~\ref{table_hybrid_result}. These results demonstrate that our model consistently performs at a top-tier level across several metrics. Notably, our model relies solely on location information from traffic agents for prediction, highlighting its robustness. It's worth mentioning that different methods in the current task employ varying data features. For example, S2TNet~\cite{S2TNet} uses additional information such as agent size and heading angle, TPNet~\cite{TPNet} incorporates geographic base map images, and methods like TrafficPredict~\cite{TrafficPredict} distinguish between different classes of traffic agents and design their structures accordingly to accommodate these diverse features.  
The visual depiction of the scenarios is presented in Figure~\ref{figure-mix}, with SA-SGTN used to represent our prediction results. Additional visualization details can be found in Appendix 3. The figure illustrates that our model is capable of handling both slow-moving pedestrians and vehicles in a turning state.
}

{
Our model excels in achieving state-of-the-art results, particularly in the prediction of vehicle and bicyclist trajectories. However, it slightly underperforms in predicting pedestrian trajectories. This discrepancy can be attributed to the fact that in hybrid trajectory prediction, where traffic agents vary in size and features, it's easier for the model to overlook slower pedestrians with more random trajectories when not distinguishing between agent classes. Consequently, the model tends to focus more on vehicles and bicyclists. Furthermore, SA-SGTN and A-SGTN outperform SGTN in predicting vehicle and bicyclist trajectories, underscoring their ability to prioritize feature-rich regions while disregarding less relevant ones for these specific agent types. These findings are consistent with previous experiments, reinforcing the crucial role of attention mechanisms, including sparse attention, in vehicle and bicyclist trajectory prediction.
}

\section{Conclusion}
{
In this work, we propose Attention-aware Social Graph Transformer Networks for trajectory prediction. 
Our model innovatively generates pseudo-images from the Spatio-temporal Graph through a specially designed stacking and interception method. 
We develope an attention-aware module that proves advantageous for vehicles compared to pedestrians in hybrid trajectory prediction scenarios.
After that, we capture the Spatio-temporal information through an adjacency matrix, use Spatio-temporal convolution for aggregation, and then perform temporal extrapolation through Transformer.
Altogether, our model combines the advantages of Graphs and Transformer to complete multi-modal trajectory prediction.
Our experimental results on the benchmark show that our model achieves state-of-the-art performance on most metrics, with the contribution of each module analyzed through ablation experiments and visualization.
The reduction in the final displacement error of pedestrian and vehicle trajectory prediction illustrates our model's more long-time predictive capability.
}



\ifCLASSOPTIONcaptionsoff
  \newpage
\fi

\bibliographystyle{IEEEtran}
\bibliography{IEEEabrv, main}

\begin{thebibliography}{10}
\providecommand{\url}[1]{#1}
\csname url@samestyle\endcsname
\providecommand{\newblock}{\relax}
\providecommand{\bibinfo}[2]{#2}
\providecommand{\BIBentrySTDinterwordspacing}{\spaceskip=0pt\relax}
\providecommand{\BIBentryALTinterwordstretchfactor}{4}
\providecommand{\BIBentryALTinterwordspacing}{\spaceskip=\fontdimen2\font plus
\BIBentryALTinterwordstretchfactor\fontdimen3\font minus \fontdimen4\font\relax}
\providecommand{\BIBforeignlanguage}[2]{{%
\expandafter\ifx\csname l@#1\endcsname\relax
\typeout{** WARNING: IEEEtran.bst: No hyphenation pattern has been}%
\typeout{** loaded for the language `#1'. Using the pattern for}%
\typeout{** the default language instead.}%
\else
\language=\csname l@#1\endcsname
\fi
#2}}
\providecommand{\BIBdecl}{\relax}
\BIBdecl

\bibitem{flow2}
A.~Baggag, S.~Abbar, A.~Sharma, T.~Zanouda, A.~Al{-}Homaid, A.~Mohan, and J.~Srivastava, ``Learning spatiotemporal latent factors of traffic via regularized tensor factorization: Imputing missing values and forecasting,'' \emph{{IEEE} Trans. Knowl. Data Eng.}, vol.~33, no.~6, pp. 2573--2587, 2021.

\bibitem{flow3}
Z.~Pan, W.~Zhang, Y.~Liang, W.~Zhang, Y.~Yu, J.~Zhang, and Y.~Zheng, ``Spatio-temporal meta learning for urban traffic prediction,'' \emph{{IEEE} Trans. Knowl. Data Eng.}, vol.~34, no.~3, pp. 1462--1476, 2022.

\bibitem{event1}
Z.~Yang, H.~Sun, J.~Huang, Z.~Sun, H.~Xiong, S.~Qiao, Z.~Guan, and X.~Jia, ``An efficient destination prediction approach based on future trajectory prediction and transition matrix optimization,'' \emph{{IEEE} Trans. Knowl. Data Eng.}, vol.~32, no.~2, pp. 203--217, 2020.

\bibitem{event2}
A.~V. Khezerlou, X.~Zhou, L.~Tong, Y.~Li, and J.~Luo, ``Forecasting gathering events through trajectory destination prediction: {A} dynamic hybrid model,'' \emph{{IEEE} Trans. Knowl. Data Eng.}, vol.~33, no.~3, pp. 991--1004, 2021.

\bibitem{Human_survey}
A.~Rudenko, L.~Palmieri, M.~Herman, K.~M. Kitani, D.~M. Gavrila, and K.~O. Arras, ``Human motion trajectory prediction: a survey,'' \emph{Int. J. Robotics Res.}, vol.~39, no.~8, 2020.

\bibitem{TrafficPredict}
Y.~Ma, X.~Zhu, S.~Zhang, R.~Yang, W.~Wang, and D.~Manocha, ``Trafficpredict: Trajectory prediction for heterogeneous traffic-agents,'' in \emph{{AAAI} 2019}, pp. 6120--6127.

\bibitem{Crowd-Robot_Interaction}
C.~Chen, Y.~Liu, S.~Kreiss, and A.~Alahi, ``Crowd-robot interaction: Crowd-aware robot navigation with attention-based deep reinforcement learning,'' in \emph{{ICRA} 2019}, pp. 6015--6022.

\bibitem{DESIRE}
N.~Lee, W.~Choi, P.~Vernaza, C.~B. Choy, P.~H.~S. Torr, and M.~Chandraker, ``{DESIRE:} distant future prediction in dynamic scenes with interacting agents,'' in \emph{{CVPR} 2017}, pp. 2165--2174.

\bibitem{YOLOv5}
Y.~Zhang and Y.~Zhou, ``Yolov5 based pedestrian safety detection in underground coal mines,'' in \emph{{ROBIO} 2021}, pp. 1700--1705.

\bibitem{Autonomous}
H.~Li, A.~V. Savkin, and B.~Vucetic, ``Autonomous area exploration and mapping in underground mine environments by unmanned aerial vehicles,'' \emph{Robotica}, vol.~38, no.~3, pp. 442--456, 2020.

\bibitem{Social_LSTM}
A.~Alahi, K.~Goel, V.~Ramanathan, A.~Robicquet, L.~Fei{-}Fei, and S.~Savarese, ``Social {LSTM:} human trajectory prediction in crowded spaces,'' in \emph{{CVPR} 2016,}, pp. 961--971.

\bibitem{Trajectron++}
T.~Salzmann, B.~Ivanovic, P.~Chakravarty, and M.~Pavone, ``Trajectron++: Dynamically-feasible trajectory forecasting with heterogeneous data,'' in \emph{{ECCV} 2020}, pp. 683--700.

\bibitem{GRIP}
X.~Li, X.~Ying, and M.~C. Chuah, ``{GRIP:} graph-based interaction-aware trajectory prediction,'' in \emph{{ITSC} 2019}, pp. 3960--3966.

\bibitem{cslstm}
N.~Deo and M.~M. Trivedi, ``Convolutional social pooling for vehicle trajectory prediction,'' in \emph{{CVPR} Workshops 2018}, pp. 1468--1476.

\bibitem{S2TNet}
W.~Chen, F.~Wang, and H.~Sun, ``S2tnet: Spatio-temporal transformer networks for trajectory prediction in autonomous driving,'' in \emph{{ACML} 2021}, pp. 454--469.

\bibitem{Trajectron}
B.~Ivanovic and M.~Pavone, ``The trajectron: Probabilistic multi-agent trajectory modeling with dynamic spatiotemporal graphs,'' in \emph{{ICCV} 2019}, pp. 2375--2384.

\bibitem{Social_GAN}
A.~Gupta, J.~Johnson, L.~Fei{-}Fei, S.~Savarese, and A.~Alahi, ``Social {GAN:} socially acceptable trajectories with generative adversarial networks,'' in \emph{{CVPR} 2018}, pp. 2255--2264.

\bibitem{Social_force}
D.~Helbing and P.~Moln\'ar, ``Social force model for pedestrian dynamics,'' \emph{Phys. Rev. E}, vol.~51, pp. 4282--4286, May 1995.

\bibitem{Gaussian_Process}
J.~M. Wang, D.~J. Fleet, and A.~Hertzmann, ``Gaussian process dynamical models for human motion,'' \emph{{IEEE} Trans. Pattern Anal. Mach. Intell.}, vol.~30, no.~2, pp. 283--298.

\bibitem{LSTM}
S.~Hochreiter and J.~Schmidhuber, ``Long short-term memory,'' \emph{Neural Comput.}, vol.~9, no.~8, pp. 1735--1780, 1997.

\bibitem{Attention_is_All}
A.~Vaswani, N.~Shazeer, N.~Parmar, J.~Uszkoreit, L.~Jones, A.~N. Gomez, L.~Kaiser, and I.~Polosukhin, ``Attention is all you need,'' in \emph{{NeurIPS} 2017}, pp. 5998--6008.

\bibitem{BERT}
J.~Devlin, M.~Chang, K.~Lee, and K.~Toutanova, ``{BERT:} pre-training of deep bidirectional transformers for language understanding,'' in \emph{{NAACL-HLT} 2019}, pp. 4171--4186.

\bibitem{flow1}
D.~A. Tedjopurnomo, Z.~Bao, B.~Zheng, F.~M. Choudhury, and A.~K. Qin, ``A survey on modern deep neural network for traffic prediction: Trends, methods and challenges,'' \emph{{IEEE} Trans. Knowl. Data Eng.}, vol.~34, no.~4, pp. 1544--1561, 2022.

\bibitem{yaoliu}
Y.~Liu, L.~Yao, B.~Li, X.~Wang, and C.~Sammut, ``Social graph transformer networks for pedestrian trajectory prediction in complex social scenarios,'' in \emph{CIKM, 2022}, pp. 1339--1349.

\bibitem{STAR}
C.~Yu, X.~Ma, J.~Ren, H.~Zhao, and S.~Yi, ``Spatio-temporal graph transformer networks for pedestrian trajectory prediction,'' in \emph{{ECCV} 2020}, pp. 507--523.

\bibitem{SR-LSTM}
P.~Zhang, W.~Ouyang, P.~Zhang, J.~Xue, and N.~Zheng, ``{SR-LSTM:} state refinement for {LSTM} towards pedestrian trajectory prediction,'' in \emph{{CVPR} 2019}, pp. 12\,085--12\,094.

\bibitem{SoPhie}
A.~Sadeghian, V.~Kosaraju, A.~Sadeghian, N.~Hirose, H.~Rezatofighi, and S.~Savarese, ``Sophie: An attentive {GAN} for predicting paths compliant to social and physical constraints,'' in \emph{{CVPR} 2019}, pp. 1349--1358.

\bibitem{Transformer_for_Trajectory}
F.~Giuliari, I.~Hasan, M.~Cristani, and F.~Galasso, ``Transformer networks for trajectory forecasting,'' in \emph{{ICPR} 2020}, pp. 10\,335--10\,342.

\bibitem{End-to-end_Transformer}
L.~L. Li, B.~Yang, M.~Liang, W.~Zeng, M.~Ren, S.~Segal, and R.~Urtasun, ``End-to-end contextual perception and prediction with interaction transformer,'' in \emph{{IROS} 2020}, pp. 5784--5791.

\bibitem{mmTransformer}
Y.~Liu, J.~Zhang, L.~Fang, Q.~Jiang, and B.~Zhou, ``Multimodal motion prediction with stacked transformers,'' in \emph{{CVPR} 2021}, pp. 7577--7586.

\bibitem{Social-BiGAT}
V.~Kosaraju, A.~Sadeghian, R.~Mart{\'{\i}}n{-}Mart{\'{\i}}n, I.~D. Reid, H.~Rezatofighi, and S.~Savarese, ``Social-bigat: Multimodal trajectory forecasting using bicycle-gan and graph attention networks,'' in \emph{NeurIPS 2019}, pp. 137--146.

\bibitem{TPCN}
M.~Ye, T.~Cao, and Q.~Chen, ``{TPCN:} temporal point cloud networks for motion forecasting,'' in \emph{{CVPR} 2021}, pp. 11\,318--11\,327.

\bibitem{Social-STGCNN}
A.~A. Mohamed, K.~Qian, M.~Elhoseiny, and C.~G. Claudel, ``Social-stgcnn: {A} social spatio-temporal graph convolutional neural network for human trajectory prediction,'' in \emph{{CVPR} 2020}, pp. 14\,412--14\,420.

\bibitem{ST-GCNN}
S.~Yan, Y.~Xiong, and D.~Lin, ``Spatial temporal graph convolutional networks for skeleton-based action recognition,'' in \emph{{AAAI} 2018}, pp. 7444--7452.

\bibitem{RCNN}
R.~B. Girshick, J.~Donahue, T.~Darrell, and J.~Malik, ``Rich feature hierarchies for accurate object detection and semantic segmentation,'' in \emph{{CVPR} 2014}, pp. 580--587.

\bibitem{graph-cnn}
T.~N. Kipf and M.~Welling, ``Semi-supervised classification with graph convolutional networks,'' in \emph{{ICLR} 2017}.

\bibitem{flow4}
S.~Guo, Y.~Lin, H.~Wan, X.~Li, and G.~Cong, ``Learning dynamics and heterogeneity of spatial-temporal graph data for traffic forecasting,'' \emph{{IEEE} Trans. Knowl. Data Eng.}, vol.~34, no.~11, pp. 5415--5428, 2022.

\bibitem{nce}
Y.~Liu, Q.~Yan, and A.~Alahi, ``Social {NCE:} contrastive learning of socially-aware motion representations,'' in \emph{{ICCV} 2021}, pp. 15\,098--15\,109.

\bibitem{ETH}
S.~Pellegrini, A.~Ess, K.~Schindler, and L.~V. Gool, ``You'll never walk alone: Modeling social behavior for multi-target tracking,'' in \emph{{ICCV} 2009}, pp. 261--268.

\bibitem{UCY}
L.~Leal{-}Taix{\'{e}}, M.~Fenzi, A.~Kuznetsova, B.~Rosenhahn, and S.~Savarese, ``Learning an image-based motion context for multiple people tracking,'' in \emph{{CVPR} 2014}, pp. 3542--3549.

\bibitem{Social_Att}
A.~Vemula, K.~Muelling, and J.~Oh, ``Social attention: Modeling attention in human crowds,'' in \emph{{ICRA} 2018}, pp. 1--7.

\bibitem{DMRGCN}
I.~Bae and H.~Jeon, ``Disentangled multi-relational graph convolutional network for pedestrian trajectory prediction,'' in \emph{{AAAI} 2021}, pp. 911--919.

\bibitem{generativeSCAN}
J.~Sekhon and C.~H. Fleming, ``{SCAN:} {A} spatial context attentive network for joint multi-agent intent prediction,'' in \emph{{AAAI} 2021}, pp. 6119--6127.

\bibitem{80}
J.~Colyar and J.~Halkias, ``Us highway i-80 dataset: Tech. rep. fhwa-hrt-06-137,'' Tech. Rep., 2006.

\bibitem{101}
------, ``Us highway 101 dataset,'' \emph{Federal Highway Administration, Tech. Rep. FHWA-HRT-07-030}, pp. 27--69, 2007.

\bibitem{cslstm-pre}
N.~Deo, A.~Rangesh, and M.~M. Trivedi, ``How would surround vehicles move? {A} unified framework for maneuver classification and motion prediction,'' \emph{{IEEE} Trans. Intell. Veh.}, vol.~3, no.~2, pp. 129--140, 2018.

\bibitem{gru-compare}
A.~Kuefler, J.~Morton, T.~A. Wheeler, and M.~J. Kochenderfer, ``Imitating driver behavior with generative adversarial networks,'' in \emph{{IV} 2017}.\hskip 1em plus 0.5em minus 0.4em\relax {IEEE}, 2017, pp. 204--211.

\bibitem{StarNet}
Y.~Zhu, D.~Qian, D.~Ren, and H.~Xia, ``Starnet: Pedestrian trajectory prediction using deep neural network in star topology,'' in \emph{{IROS} 2019}, pp. 8075--8080.

\bibitem{TPNet}
L.~Fang, Q.~Jiang, J.~Shi, and B.~Zhou, ``Tpnet: Trajectory proposal network for motion prediction,'' in \emph{{CVPR} 2020}, pp. 6796--6805.

\bibitem{GRIP++}
X.~Li, X.~Ying, and M.~C. Chuah, ``Grip++: Enhanced graph-based interaction-aware trajectory prediction for autonomous driving,'' \emph{arXiv preprint arXiv:1907.07792}, 2019.

\end{thebibliography}

\end{document}